\documentclass[letterpaper]{article} % DO NOT CHANGE THIS
\usepackage{aaai23}  % DO NOT CHANGE THIS
\usepackage{times}  % DO NOT CHANGE THIS
\usepackage{helvet}  % DO NOT CHANGE THIS
\usepackage{courier}  % DO NOT CHANGE THIS
\usepackage[hyphens]{url}  % DO NOT CHANGE THIS
\usepackage{graphicx} % DO NOT CHANGE THIS
\usepackage{natbib}  % DO NOT CHANGE THIS AND DO NOT ADD ANY OPTIONS TO IT
\usepackage{caption} % DO NOT CHANGE THIS AND DO NOT ADD ANY OPTIONS TO IT
\usepackage[utf8]{inputenc} % allow utf-8 input
\usepackage[T1]{fontenc}    % use 8-bit T1 fonts
\usepackage{url}            % simple URL typesetting
\usepackage{booktabs}       % professional-quality tables
\usepackage{amsfonts}       % blackboard math symbols
\usepackage{nicefrac}       % compact symbols for 1/2, etc.
\usepackage{microtype}      % microtypography
\usepackage{xcolor}         % colors

\usepackage{amsmath}
\usepackage{amssymb}
\usepackage[output-decimal-marker=\text{.}, range-phrase = \text{--}]{siunitx}
\usepackage{algorithm}
\usepackage{algorithmic}

\usepackage{graphicx}
\usepackage{overpic}
\usepackage{wrapfig}

\DeclareMathOperator*{\dd}{d}
\DeclareMathOperator*{\argmin}{arg\,min}
\DeclareMathOperator*{\argmax}{arg\,max}

\newtheorem{prop}{Proposition}
\newcommand{\code}[1]{\texttt{#1}}

\DeclareCaptionStyle{ruled}{labelfont=normalfont,labelsep=colon,strut=off} % DO NOT CHANGE THIS

\title{An Operator Theoretic Approach for Analyzing Sequence Neural Networks}
\author {
    Ilan Naiman,
    Omri Azencot
}
\affiliations {
    Department of Computer Science,\\
    Ben-Gurion University of the Negev, Beer Sheva, Israel\\
    naimani@post.bgu.ac.il, azencot@cs.bgu.ac.il
}
\begin{document}

\maketitle

% \vskip 0.3in

% \vspace{-3mm}
\begin{abstract}

Analyzing the inner mechanisms of deep neural networks is a fundamental task in machine learning. Existing work provides limited analysis or it depends on local theories, such as fixed-point analysis. In contrast, we propose to analyze trained neural networks using an operator theoretic approach which is rooted in Koopman theory, the Koopman Analysis of Neural Networks (\code{KANN}). Key to our method is the Koopman operator, which is a linear object that globally represents the dominant behavior of the network dynamics. The linearity of the Koopman operator facilitates analysis via its eigenvectors and eigenvalues. Our method reveals that the latter eigendecomposition holds semantic information related to the neural network inner workings. For instance,  the eigenvectors highlight positive and negative $n$-grams in the sentiments analysis task; similarly, the eigenvectors capture the salient features of healthy heart beat signals in the ECG classification problem.

\end{abstract}

\section{Introduction}

Understanding the inner workings of predictive models is an essential requirement in many fields across science and engineering. This need is even more important nowadays with the emergence of neural networks whose visualization and interpretation is inherently challenging. Indeed, modern computational neural models often lack a commonly accepted knowledge regarding their governing mathematical principles. Consequently, while deep neural networks may achieve remarkable results on various complex tasks, explaining their underlying decision mechanisms remains a challenge. The goal of this paper is to help bridge this gap by proposing a new framework for the approximation, reasoning, and understanding of sequence neural models.

Sequence models are designed to handle time series data originating from images, text, audio, and other sources of information. One approach to analyzing sequence neural networks is through the theory and practice of dynamical systems \citep{doya1993bifurcations, pascanu2013difficulty}. For instance, the temporal asymptotic behavior of a dynamical system can be described using the local analysis of its attractor states~\citep{strogatz2018nonlinear}. Similarly, recurrent models have been investigated in the neighborhood of their fixed points \citep{sussillo2013opening}, leading to work that interprets trained RNNs for tasks such as sentiment analysis \citep{maheswaranathan2019reverse}. 

However, the local nature of these methods is a limiting factor which may lead to inconsistent results. Specifically, their approach is based on fixed-point analysis which allows to study the dynamical system in the neighborhood of a fixed-point. In contrast, our approach is global---it does not depend on a set of fixed-points, and it facilitates the exploration of the dynamics near and further away from fixed points.

Over the past few years, a family of data-driven methods was developed, allowing to analyze complex dynamical systems based on Koopman theory (\citeyear{koopman1931hamiltonian}). These methods exploit a novel observation by which nonlinear systems may be globally encoded using infinite-dimensional but \emph{linear} Koopman operators. In practice, Koopman-based approaches are lossy as they compute a low-rank approximation of the full operator. Nevertheless, it has been shown in the fluid dynamics~\citep{mezic2005spectral, azencot2020forecasting} and geometry processing~\citep{ovsjanikov2012functional, sharmaO2020weakly} communities that the dominant features of general nonlinear dynamical systems can be captured via a single matrix per system, allowing e.g., to forecast dynamics~\citep{azencot2019consistent,cohen2021modes}, and to align time series data~\citep{rahamim2021aligning}. Thus, we pose the following research question: can we design and employ a Koopman-based approach to analyze and develop a fundamental understanding of deep neural models?

Given a trained sequence neural network and a procedure to extract its hidden states, our Koopman-based method generates a moderate size matrix which faithfully describes the dynamics in the latent space. Unlike existing work, our approach is global and independent of a particular latent sample, and thus it can be virtually applied to any hidden state. A key advantage of our framework is that we can directly employ linear analysis tools on the approximate Koopman operator to reason about the associated neural network. In particular, we show that the eigenvectors and eigenvalues of the Koopman matrix are instrumental for understanding the decision mechanisms of the model. For instance, we show in our results that the dominant eigenvectors carry crucial \emph{semantic knowledge} related to the problem at hand. Koopman-based methods such as ours are backed by rich theory and practice, allowing us to exploit the recent advances in Koopman inspired techniques for the purpose of developing a comprehensive understanding of sequence neural networks. Thus, the \textbf{key contribution} in this work is the novel application of Koopman-based methods for understanding sequential models, and the extraction of high-level interpretable and insightful understandings on the trained networks.

We focus our investigation on two learning tasks: sentiment analysis and electrocardiogram (ECG) classification. We will identify four eigenvectors in the sentiment analysis model whose roles are to highlight: positive words, negative words, positive pairs (e.g., ``not bad''), and negative pairs.  In addition, we demonstrate that the eigenvectors in the ECG classification task naturally identify dominant features in normal beat signals and encode them. Specifically, we show that four Koopman eigenvectors accurately capture the local extrema points of normal beat signals. These extrema points are fundamental in deciding whether a signal is normal or anomalous. Our results reinforce that the network indeed learns a robust representation of normal beat signals. Then, we will verify that the main components of the nonlinear network dynamics can be described using our Koopman matrices by measuring the difference in accuracy results, and the relative error in predicted states. Further, we provide additional results and comparisons in the supplementary material. Given the versatility of our framework and its ease of use, we advocate its utility in the analysis and understanding of neural networks, and we believe it may also affect the design and training of deep models in the future. Code to reproduce our results is available at \url{https://github.com/azencot-group/KANN}.

\section{Related Work}

\paragraph{Recurrent Neural Networks and Dynamical Systems.} 

Fully connected recurrent neural networks are universal approximators of arbitrary dynamical systems \citep{doya1993universality}. Unfortunately, RNNs are well-known to be difficult to train~\citep{bengio1993problem,pascanu2013difficulty}, and several methods use a dynamical systems view to improve training via gradient clipping \citep{pascanu2013difficulty}, and constraining weights \citep{erichson2020lipschitz, azencot2021differential}, among other approaches. 
Overall, it is clear that dynamical systems are fundamental in investigating and developing recurrent networks.

\paragraph{Understanding RNN.} 

Establishing a deeper understanding of recurrent networks is a long standing challenge in machine learning. To this end,
\citet{karpathy2015visualizing} follow the outputs of the model to identify units which track brackets, line lengths, and quotes. 
Recently, \citet{chefer2020transformer} proposed an approach for computing relevance scores of transformer networks. Perhaps mostly related to our approach is the analysis of recurrent models around their fixed points~\citep{sussillo2013opening}.
This approach revealed low-dimensional attractors in the sentiment analysis task~\citep{maheswaranathan2019reverse}, which allowed to deduce simple explanations of the decision mechanisms. Our work generalizes the approach of \citet{sussillo2013opening} in that it yields global results about the dynamics, and it introduces several novel features. We provide a more detailed comparison between our method and theirs in Sec.~\ref{sec:results}.

% Koopman-based methods
\paragraph{Koopman-based Neural Networks.}

Recently, several techniques that combine neural networks and Koopman theory were proposed, mostly in the context of \emph{predicting} nonlinear dynamics. For example, \citet{takeishi2017learning, morton2018deep} optimize the residual sum of squares of the predictions the operator makes, \citet{lusch2018deep, erichson2019physics, azencot2020forecasting} design dynamic autoencoders whose central component is linear and may be structured, \citet{li2020learning} employ graph networks, and \citet{mardt2018vampnets} use a variational approach on Markov processes. A recent line of work aims at exploiting tools from Koopman theory to analyze and improve the training process of neural networks~\citep{dietrich2020koopman, dogra2020optimizing, manojlovic2020applications}. To the best of our knowledge, our work is first to employ a Koopman-based method towards the analysis and understanding of trained neural networks.

% \clearpage
\section{Method}
\label{sec:mtd}

% neural models in consideration
In what follows, we present our method for analyzing and understanding sequence neural models. Importantly, while we mostly discuss and experiment with recurrent neural networks, our approach is quite general and applicable to any model whose inner representation is a time series. We consider neural models that take input instances $x_t \in \mathbb{R}^m$ at time $t$ and compute
\begin{align} \label{eq:neural_dynamics}
    h_t = F(h_{t-1},\, x_t) \ , \quad t=1,2,... \ ,
\end{align}
where $h_t \in \mathbb{R}^k$ is a (hidden) state that represents the latent dynamics, and $F$ is some nonlinear function that pushes states through time. In our analysis, we use only the hidden states set and discard the time series input. Thus, our method is a ``white-box'' approach as we assume access to $\{ h_t \}$, which is typically possible in most day-to-day scenarios. Importantly, all recurrent models including vanilla RNN \citep{elman1990finding}, LSTM~\citep{hochreiter1997long}, and GRU \citep{cho2014learning}, as well as Attention Models~\citep{bahdanau14neural, vaswani2017attention}, and Residual neural networks~\citep{he2016deep} exhibit the structure of Eq.~\eqref{eq:neural_dynamics}.

% Koopman theory, Koopman operator, states vs. observables
\subsection{Essentials of Koopman theory}
\label{subsec:theory}

Our approach is based on \citet{koopman1931hamiltonian} theory which was developed for dynamical systems. The key observation of Koopman was that a finite-dimensional nonlinear dynamics can be fully represented using an infinite-dimensional but \emph{linear} operator. While the theoretical background is essential for developing a deep understanding of Koopman-based approaches, the practical aspects are more important to this work. Thus, we briefly recall the definition of the \emph{Koopman operator}, and we refer the reader to other, comprehensive works on the subject~\citep{singh1993composition, eisner2015operator}. Formally, we assume a discrete-time dynamical system $\varphi$ acting on a compact, inner-product space $\mathcal{M} \subset \mathbb{R}^m$,
\begin{equation}
    z_{t+1} = \varphi(z_t) \ , \quad z_t \in \mathcal{M} \ , \quad t=1,2,... \ ,
\end{equation}
where $t$ is an integer index representing discrete time. The dynamics $\varphi$ induces a linear operator $\mathcal{K}_\varphi$ which we call the Koopman operator, and it is given by
\begin{equation} \label{eq:koopman_op}
    \mathcal{K}_\varphi f(z_t) := f(z_{t+1}) = f \circ \varphi(z_t) \ ,
\end{equation}
where $f : \mathcal{M} \rightarrow \mathbb{R}$ is a scalar function in a bounded inner product space $\mathcal{F}$. It is easy to show that $\mathcal{K}_\varphi$ is linear due to the linearity of composition, i.e., given $\alpha, \beta \in \mathbb{R}$ and $f, g \in \mathcal{F}$, we obtain that $\mathcal{K}_\varphi (\alpha f + \beta g) = (\alpha f + \beta g) \circ \varphi = \alpha f \circ \varphi + \beta g \circ \varphi = \alpha \mathcal{K}_\varphi(f) + \beta \mathcal{K}_\varphi(g)$. We emphasize that while $\varphi$ describes the system evolution, $\mathcal{K}_\varphi$ is a transformation on the space of \emph{functions}. From a practical viewpoint, these functions may be interpreted as observations of the system, such as velocity, sea level, temperature, or hidden states.

To justify our use of Koopman theory and practice in the context of neural networks, we propose the following. We interpret the input sequence $\{ x_t \}$ as governed by some complex and unknown dynamics $\varphi$, i.e., $x_{t+1} = \varphi(x_t)$ for every $t$. We emphasize that $\varphi$ is different from $F$ in Eq.~\eqref{eq:neural_dynamics} by its definition of domain and range. Then, the hidden states $h_t$ are finite samplings of observations of the system, namely, $h_t \approx f_t$ where $f_t:\mathcal{M} \rightarrow \mathbb{R}$ is the true observation. For instance, $f_t$ may be the smooth function $\cos(t z)$, whereas $h_t \in \mathbb{R}^k$ is its sampling at a finite set of points $\{ z_1,\dots,z_k \}$. It follows that $\{ h_t \}$ is subject to an approximate Koopman representation. However, a fundamental challenge in facilitating Koopman theory in practice is the infinite-dimensionality of $\mathcal{K}_\varphi$. Recently, several data-driven methods were developed to produce a better approximate $\mathcal{K}_\varphi$ using a moderate-size matrix $C$~\citep{schmid2010dynamic, ovsjanikov2012functional}. In particular, Koopman-based approaches have been proven instrumental in the analysis of fluid dynamics data~\citep{brunton2021modern}, and for computing complex nonrigid isometric maps between shapes~\citep{ovsjanikov2016computing}. Motivated by these empirical examples and their success, we will compute in this work approximate Koopman operator matrices $C$ such that they encode the evolution of latent states $\{ h_t \}$.

% DMD/FMAPS: choice of basis, least squares
\subsection{A Koopman-based method}
\label{subsec:mtd}
We denote by $H \in \mathbb{R}^{s \times n \times k}$ a tensor of hidden state sequences, where $s$ is the batch size, $n$ is the sequence length and $k$ is the hidden dimension. The method we employ for computing the matrix $C$ follows two simple steps: 1. Represent the states using a basis $B$, and denote the resulting collection of spectral coefficients by $\tilde{H}$. 2. Find the best linear transformation $C$ which maps $\tilde{H}_t$ to $\tilde{H}_{t+1}$ in the spectral domain, where $\tilde{H}_\tau \in \mathbb{R}^{s \times k}$ denotes the tensor of coefficients from $\tilde{H}$ at time $\tau$. To give a specific example of the general procedure we just described, we can choose the principal components $b_j, j=1,2,...$ of the truncated \code{SVD} of the states $H$ to be the basis in the first step. Then, the resulting basis elements are orthonormal, i.e., $B^T B = \mathrm{Id}$, where $B = (b_j)$ is the matrix of basis elements organized in its columns, and $\mathrm{Id}$ is the identity matrix. The matrix $C$ is obtained by solving the following least squares minimization
\begin{align} 
    & C := \argmin_{\tilde{C}} \sum_{t=1}^{n-1}{\left| \tilde{H}_t \cdot \tilde{C} - \tilde{H}_{t+1} \right|_F^2} \ , \label{eq:lstsq} \\ 
    & \tilde{H}_\tau = H_\tau \cdot B \ , \quad \forall \tau \ , \label{eq:linear_proj}
\end{align}
where $\cdot$ is matrix multiplication. We note that the above scheme is a variant of the dynamic mode decomposition~\citep{schmid2010dynamic}, and the functional maps~\citep{ovsjanikov2012functional} algorithms.

% prediction
\subsection{Koopman-based prediction}

The infinite-dimensional Koopman operator in Eq.~\eqref{eq:koopman_op} describes the evolution of observable functions subject to the dynamics $\varphi$. Similarly, our $C$ matrices allow us to predict a future hidden state $h_{t+1}$ from a given current state $h_t$ simply by multiplying $C$ with the spectral coefficients $\tilde{h}_t$. Namely, 
\begin{align} \label{eq:kann_linear_state}
    H_{t+1}^{\code{KANN}} := H_t\cdot B \cdot C \cdot B^T \ .
\end{align}
We will mostly use Eq.~\eqref{eq:kann_linear_state} to evaluate the validity of $C$ in encoding the underlying dynamics based on the differences $| H_t^{\code{KANN}} - H_t |_F^2 / | H_t |_F^2$ for every admissible $t$, see Sec.~\ref{sec:results}.

% analysis
\subsection{Koopman-based analysis}

The key advantage of Koopman theory and practice is that linear analysis tools can be directly applied to study the behavior of the underlying dynamical system. The tools we describe next form the backbone of our analysis framework, and our results are heavily based on these tools.

% eigenvectors of $C$, sepraable dynamics
\paragraph{Separable dynamics.} If $C \in \mathbb{R}^{k \times k}$ admits an eigendecomposition, then the dynamics can be represented in a fully \emph{separable} manner, where the eigenvectors of $C$ propagate along the dynamics independently of the other eigenvectors, scaled by their respective eigenvalues. Formally, we consider the eigenvalues $\lambda_j \in \mathbb{C}$ and eigenvectors $v_j \in \mathbb{C}^k$ of $C$, i.e., it holds that $C \, v_j = \lambda_j v_j$. We assume that $C$ is full-rank and thus $V = ( v_j )$ forms a basis of $\mathbb{R}^k$, and similarly, $U = V^{-1}$ is also a spanning basis. In our setting, we call the rows of $U$ the \textbf{Koopman eigenvectors}, and we represent any hidden state $h_t$ in this basis, similarly to Eq.~\eqref{eq:linear_proj}. The projection of $H$ onto $U$ reads
\begin{equation} \label{eq:eig_linear_proj}
    \hat{H}_\tau := H_\tau \cdot B \cdot V = \tilde{H}_\tau \cdot V \ .
\end{equation}
Then, re-writing Eq.~\eqref{eq:kann_linear_state} using the eigendecomposition of $C = V \cdot \Lambda \cdot U$ yields the temporal trajectory of $H_t$ via $\hat{H}_{t+1} = H_{t+1} \cdot B \cdot V \approx H_t \cdot B \cdot C \cdot V = H_t \cdot B \cdot V \cdot \Lambda  = \hat{H}_t \cdot \Lambda$, where $\Lambda$ is the diagonal matrix of eigenvalues, and the approximation is due to Eq.~\eqref{eq:kann_linear_state}. The latter derivation yields
\begin{equation}
    \hat{H}_{t+1} \approx \hat{H}_t \cdot \Lambda \ ,
\end{equation}
i.e., the linear dynamics matrix represented in the basis $U$ is simply the diagonal matrix $\Lambda$, and thus $U$ may be viewed as a ``natural'' basis for encoding the dynamics. Further, it directly follows that $\hat{H}_{t+l} \approx \hat{H}_t \cdot \Lambda^l$, that is, the number of steps forward is determined by the eigenvalues power.

\section{Results}
\label{sec:results}

In this study, we focus our exploration on the sentiment analysis and the ECG classification problems. Unless noted otherwise, we always compute $C$ using the method in Sec.~\ref{subsec:mtd}, where the basis is given by the truncated \code{SVD} modes of the input hidden states, and $C$ is the least squares estimation obtained from solving~\eqref{eq:lstsq}. We first provide our qualitative analysis, and then, we include a quantitative evaluation of \code{KANN} and its ability to encode the dynamics. In Apps.~\ref{app:results_sentiment_analysis}, \ref{app:results_sentiment_analysis_ngram}, \ref{app:results_ecg}, and \ref{app:basis_and_archs}, we provide additional results, and we show that our method is robust to the choice of basis and network architecture. Finally, we further use \code{KANN} to analyze the copy problem in App.~\ref{app:copy_task}, where our results outperform the baseline approach~ \citep{maheswaranathan2019reverse}.

\begin{figure*}[t]
  \centering
  \begin{overpic}[width=.9\linewidth]{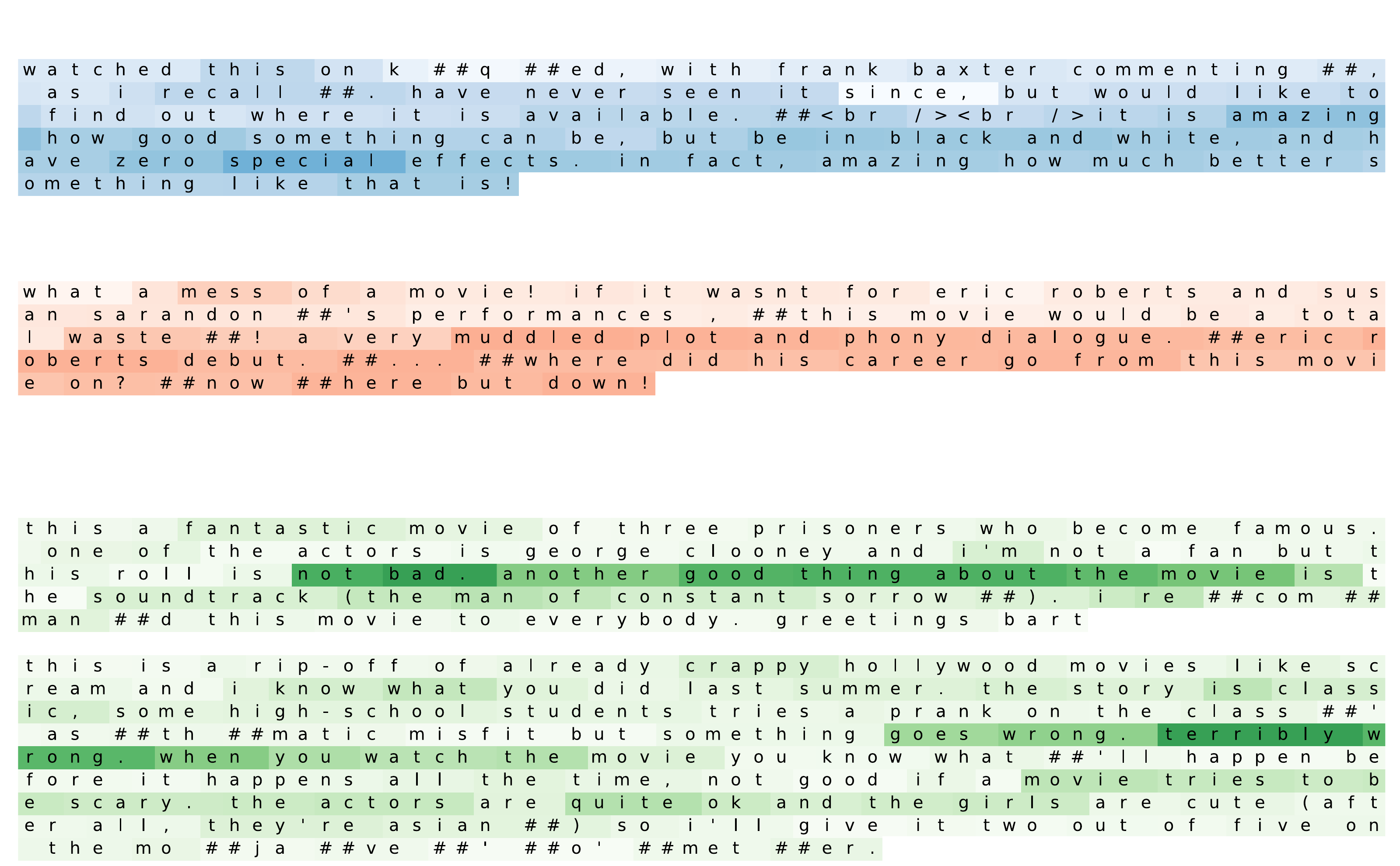} 
    \put(0,60){A positive review projected onto $\{u_1, u_2\}$:}
    \put(0,44){A negative review projected onto $\{u_3, u_4\}$:}
    \put(0,29){Two reviews with contextual information projected onto $\{u_{8}\}$:}
  \end{overpic}
  \caption{We display reviews where each word is shaded based on $\sum_j s(j, h_t)$. Projecting onto $U_{12}$ shows an increase in magnitude for several positive words (blue), whereas projecting onto $U_{34}$ shows jumps in magnitude around negative words (red). See e.g., \code{amazing, special} (blue), \code{mess, waste, muddled} (red). We also show two reviews with contextual information which is naturally highlighted due to $u_8$.}
  \label{fig:seq_proj_scale}
\end{figure*}

% explain the dynamics of sentiment analysis: four eigenvectors with "long" memory, two count positive/negative words, two other? show histogram, 
\subsection{Sentiment analysis}
\label{subsec:kann_qualitative_sa}

We begin our qualitative study by considering the sentiment analysis task which was extensively explored in~\citep{maheswaranathan2019reverse, maheswaranathan2020recurrent}. Determining the sentiment of a document is an important problem which may be viewed as a binary classification task.
%~\cite{zhang2018deep}. 
We will use the IMDB reviews dataset, 
%~\citep{maas2011learning}
and we will embed the corpus of words to obtain a vector representation of text. Given a review, the role of the network is to output whether it reflects a positive or negative opinion. Adopting the setup of \citet{maheswaranathan2019reverse}, we use a word embedding of size $128$, and a GRU recurrent layer with a hidden size of $256$. We train the model for $5$ epochs during which it reaches an accuracy of $\approx 92\%,\, 87\%,\, 87\%$ on the train, validation and test sets, respectively. For analysis, we extract a random test batch of $64$ reviews and its states $H \in \mathbb{R}^{64 \times 1000 \times 256}$, where $1000$ is the review length when padded with zeros.

One of the main results in~\citep{maheswaranathan2019reverse} was the observation that the dynamics of the network span a line attractor. That is, the hidden states of the network are dominantly attracted to a one dimensional manifold, splitting the domain into positive and negative sentiments. Additionally, \citet{maheswaranathan2020recurrent} study inputs with contextual relations (e.g., the phrase ``not bad''), and their effect on the network dynamics. Our results align with the observations in~\citep{maheswaranathan2019reverse, maheswaranathan2020recurrent}. Moreover, we generalize their results by showing that the attracting manifold is in fact of a higher dimension, and that the manifold can be decomposed to semantically understandable components using \code{KANN}. Specifically, we demonstrate that several Koopman eigenvectors are important in the dynamics, and we can link each of these eigenvectors to a semantically meaningful action. Thus, in comparison to the framework proposed in~\citep{maheswaranathan2019reverse, maheswaranathan2020recurrent}, our method naturally decomposes the latent manifold into interpretable attracting components. In addition, we provide a unified framework for reasoning and understanding by drawing conclusions directly from the separable building blocks of the latent dynamics.

Most of our results for the sentiment analysis problem are based on the eigendecomposition of $C$, its resulting eigenvalues $\{ \lambda_j \in \mathbb{C} \}$ and corresponding eigenvectors $\{ u_j \in \mathbb{C}^k \}$. For the random states batch $H$ specified above, we obtain an operator $C$ whose spectrum consists of \emph{four} eigenvalues with modulus greater than $0.99$, i.e., $| \lambda_j | > 0.99$. In comparison, \citet{maheswaranathan2019reverse} identify only a single dominant component. The values of our $\lambda_j$ read $\lambda_1 = 0.9999$. $\lambda_2 = 0.9965$, and $\lambda_{3,4} = 0.9942 \pm i0.0035$. Consequently, their respective eigenvectors have long memory horizons. Namely, if $|\lambda_j| \approx 1$, these eigenvectors carry information across long word sequences. Otherwise, if $|\lambda_j| < 1$, then its powers decay exponentially to zero.

\begin{figure*}[t]
  \centering
  \includegraphics[width=.8\linewidth]{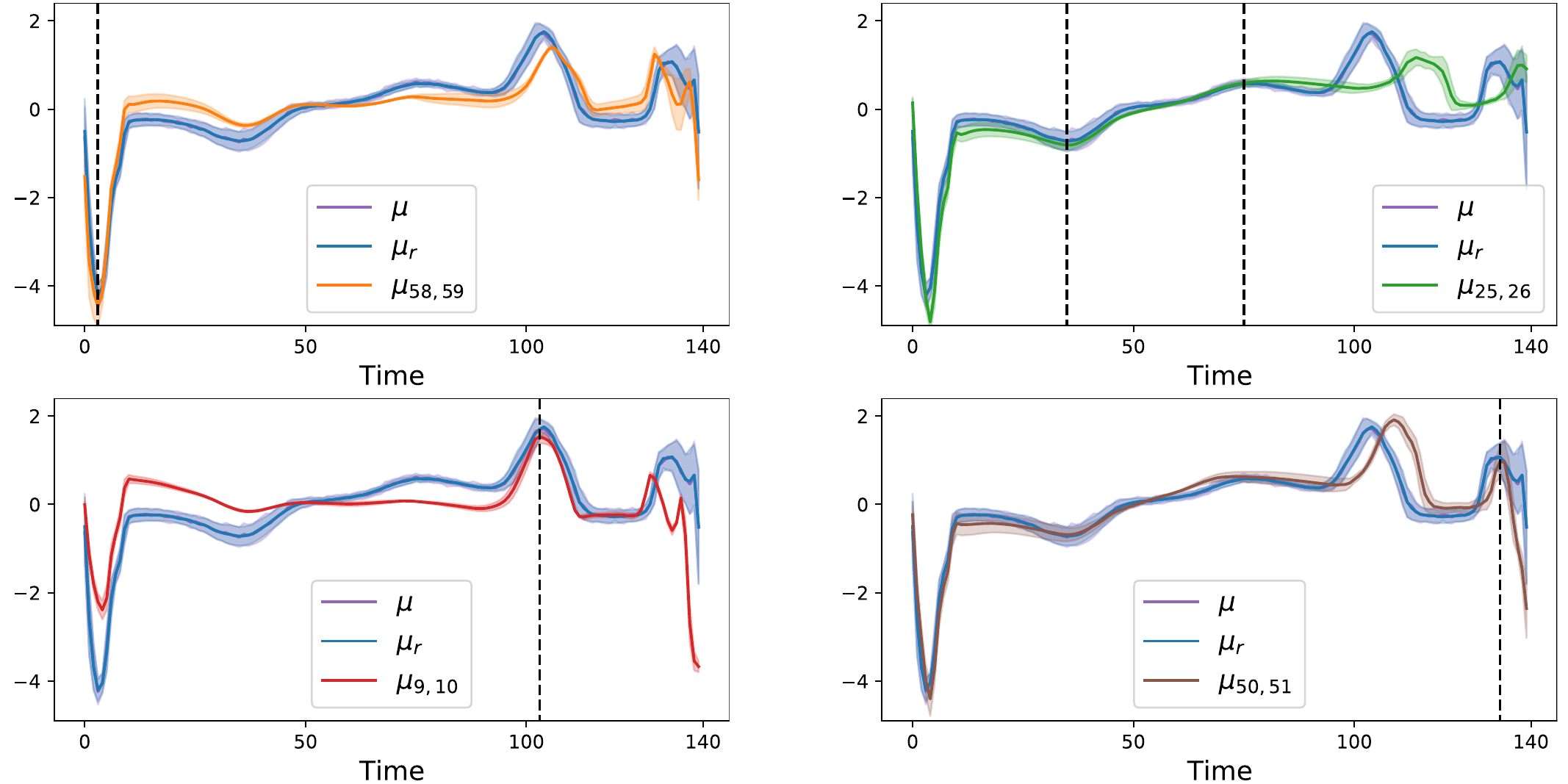}
  \caption{We show the median of reconstructions of normal beats when projected to each of the first four dominant conjugate pairs of Koopman eigenvectors. The medians (orange, green, red, brown) are plotted on top of the original signals and their reconstruction medians. The dashed black lines indicate important features of the signals which are well captured by the Koopman eigenvectors.}  
  \label{fig:ecg_eigs}
\end{figure*}

In their analysis, the authors of~\citep{maheswaranathan2019reverse} observe that the network mainly counts positive vs. negative words in a review along a line attractor. We hypothesize that in our setting, the dominant eigenvectors $\{u_1, ..., u_4\}$ are responsible for this action. To verify our hypothesis, we use the readout (linear) layer of the model to generate the logits of the state when projected to $U_{12}$ and $U_{34}$. We denote by $\tilde{y}_{12}$ and $\tilde{y}_{34}$ the logits for $\tilde{H} \cdot V_{12}$ and $\tilde{H} \cdot V_{34}$, respectively, where $V_{ij}$ denote the $i$ and $j$ columns of $V$. For the above test batch, we get perfect correspondence, i.e., $\tilde{y}_{12} < .5$ and $\tilde{y}_{34} > .5$ on all samples. In addition to encoding a certain sentiment, the Koopman eigenvectors are advantageous in comparison to a single line attractor as they allow for a direct visualization of the importance of words in a review. Specifically, we define the projection magnitude of a hidden state as follows
\begin{equation} \label{eq:proj_mag}
    s(j, h_t) := \left| \hat{h}_t(j) \right| = \left| \tilde{h}_t^T \, V_j \right| \ .
\end{equation}
We show in Fig.~\ref{fig:seq_proj_scale} two examples where the magnitude of projection onto $U_{12}$ and $U_{34}$ clearly highlights positive and negative words, respectively. In particular, as the network ``reads'' the review and identifies e.g., a negative word, it increases $s(\cdot)$. For instance, see \code{mess} and \code{muddled} in Fig.~\ref{fig:seq_proj_scale}. Importantly, there may be occurrences of positive/negative $n$-grams which are not highlighted, such as the word \code{good} in the positive example. We show in App.~\ref{app:results_sentiment_analysis} that the above results extend to the entire test set. Thus, we conclude that $\{u_1,...,u_4\}$ track positive and negative words.

In addition to $\{ u_1, ..., u_4\}$, we also want to understand how other eigenvectors affect the latent dynamics. We hypothesize that other vectors are responsible to track contextual information such as amplifiers (``extremely good'') and negations (``not bad''). We collected all reviews that include the phrases ``not bad'' and ``not good'' into a single batch, yielding a states tensor with $256$ samples. One way to check our hypothesis is to employ the former visualization using other eigenvectors. We show two such examples in Fig.~\ref{fig:seq_proj_scale} in green, where phrases such as ``not bad'', ``terribly wrong'', and ``quite ok'' are highlighted when projected onto $u_8$. We provide additional results and analysis on the identification of amplifier words and negations using \code{KANN} in App.~\ref{app:results_sentiment_analysis}. To better visualize the importance of the Koopman eigenvectors and their ordering, we projected the batch on the eigenvectors and sorted the values of Eq.~\eqref{eq:proj_mag} from high to low when summed over time and averaged over batch. In App.~\ref{app:results_sentiment_analysis_ngram}, the resulting graph shows a hierarchical behavior, where each group of eigenvectors have a different role. We consider in App.~\ref{app:results_sentiment_analysis_ngram} the general case of $n$-grams where $n>2$.

\begin{figure*}[t]
  \centering
  \includegraphics[width=.8\linewidth]{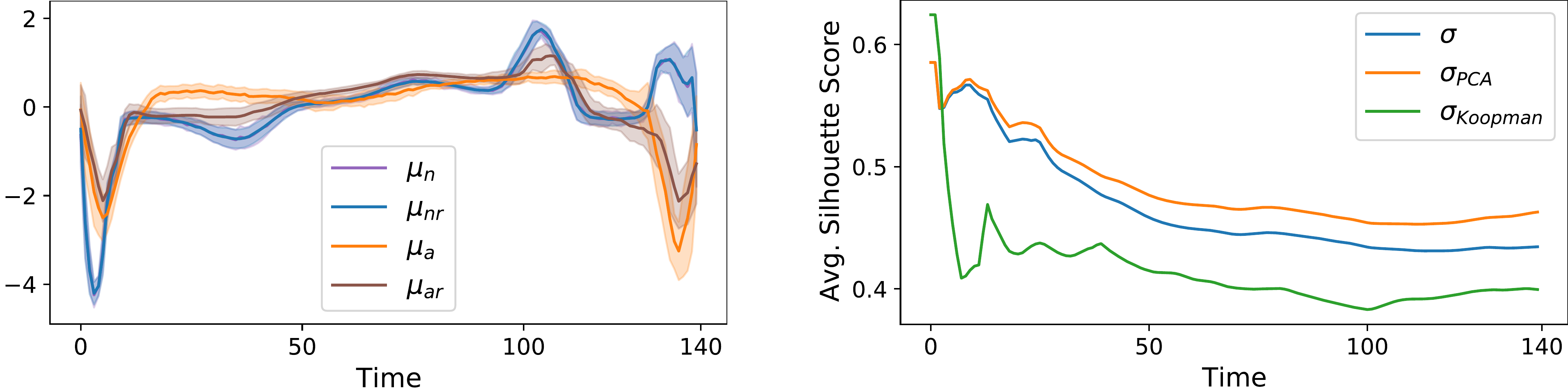}
  \caption{The network reconstructs relatively well both normal and anomalous signals (left), implying its inner representation is binary. However, the silhouette scores of the Koopman embedding (right) imply that only a single class is being learnt in practice.}  
  \label{fig:ecg_sa}
\end{figure*}

% explain the dynamics of ECG classification: 
\subsection{ECG classification}
\label{subsec:kann_qualitative_ecg}

Electrocardiogram (ECG) tests track the electrical activity in the heart, and they help detect various abnormalities in a non-invasive way. Classifying whether a beat is normal or not is a challenging task
%~\citep{zhu2019electrocardiogram} 
which lacks descriptive neural models. A common approach for solving the classification problem using neural networks trains an autoencoder model with an $L^1$ loss over the normal beats. Classification is performed by measuring the loss between the original and reconstructed signals; thus, while it is a classification task, ECG classification is solved via a \emph{regression} model. In particular, high loss values indicate anomalous beats, whereas low values are attributed to normal signals. Typically, a threshold is set during the training phase, allowing automatic classification on the test set. We fix the threshold to be $26$. Our network is composed of a single layer LSTM encoder $F_\text{enc}$ with a hidden size of $64$, and an LSTM decoder $F_\text{dec}$ with one layer as well. We use a publicly available subset of the MIT-BIH arrhythmia database~\citep{goldberger2000physiobank} for our data, named ECG5000\footnote{\url{http://timeseriesclassification.com/description.php?Dataset=ECG5000}}. This dataset includes $5000$ sample heartbeats with a sequence length of $140$. Around $60\%$ of the sequences are classified as normal and the rest are various anomalous signals. The model is trained for $150$ epochs, yielding an accuracy of $97.1\%, 97.6\%, 98.6\%$ on the train, validation and test sets, respectively.

Similarly to the sentiment analysis problem, we expect the Koopman eigenvectors to take a significant role in encoding the latent dynamics. Given that the network is generative as it is an autoencoder, we hypothesize that the eigenvectors $\{ u_j \}$ capture dominant features of normal beats. Thus, we project normal beats onto pairs of dominant eigenvectors, and decode the resulting hidden states using the decoder to study the obtained signals. For example, say $U_{58,59}$ are dominant, then we project onto the space spanned by this pair, then project back and decode. Using a test batch of $64$ \emph{normal} beats we collect the last hidden state of every sample in a matrix $H_\text{no} \in\mathbb{R}^{64 \times 64}$ and  we compute the following.
\begin{equation} \label{eq:eig_part_proj} % TODO: real part or absolute value?
    \mathbb{R}^{64 \times 140} \ni \bar{X}^{ij} = F_\text{dec}(H_\text{no} \cdot B \cdot |V_{ij} \cdot U_{ij}| \cdot B^T) \ ,
\end{equation}
where $| \cdot |$ is the element-wise modulus of complex numbers. To determine which eigenvectors are dominant, we employ Eq.~\eqref{eq:proj_mag}.

To visualize the results, we take the set of reconstructed signals $\bar{X}^{ij}$ of a particular pair $ij$, and we compute its median $\mu_{ij}(t)$ for $t \in [1,\dots,n]$. We plot these graphs in Fig.~\ref{fig:ecg_eigs} using colors orange (pair \numrange{58}{59}), green (pair \numrange{25}{26}), red (pair \numrange{9}{10}), and brown (pair \numrange{50}{51}). In addition, the original signals' median $\mu$ (purple) and the median of the signals reconstruction $\mu_r$ (blue) are shown in each of the subplots for comparison. Indeed, $\mu$ and $\mu_r$ are almost indistinguishable, and can be differentiated only when zooming in. Each median graph is wrapped in its median absolute deviation envelope. We preferred median-based quantities over the common mean and standard deviation since the latter produce cluttered plots in our setting due to outliers. To understand which time steps are important, we investigated (manually) the differences between normal and anomalous signals as they appear in the dataset. Namely, we classified the segments of the normal signals, which differed from the related parts in the anomalous signals. These segments are common across normal signals, and thus, more sensitive in terms of reconstruction error, making them "salient". The plots in Fig.~\ref{fig:ecg_eigs} clearly show that each conjugate pair captures a different feature of the time series as marked by the vertical dashed lines. Specifically, $\mu_{58,59}$ captures the minimum around $t=3$, and $\mu_{25,26}$ encodes the part of the signal in $t \in [35, 75]$. Moreover, $\mu_{9, 10}$ attains the maximum at $t=103$, and $\mu_{50,51}$ is approximating the lower peak at $t=133$ and we consider these $t$ values to be the salient features. Importantly, the other Koopman eigenvectors beyond the ones we consider above are less important in the reconstruction, and are mostly helpful in fixing minor variations. Finally, we provide a similar computation in App.~\ref{app:results_ecg} using the dominant \code{PCA} modes and \code{KernelPCA} eigenvectors, where we show that \code{PCA} components and \code{KernelPCA} eigenvectors are not useful in identifying the dominant features of beat signals. Also, we provide a quantitative comparison between the methods.
\begin{figure*}[t]
  \centering
  \includegraphics[width=.75\linewidth]{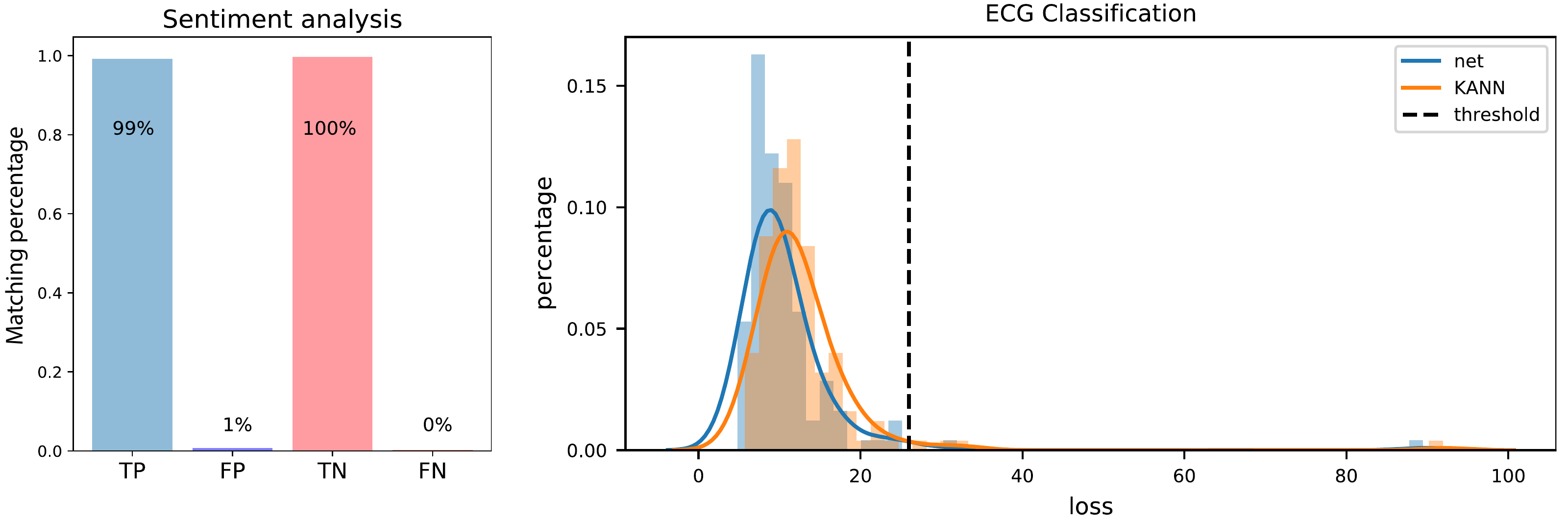}
%   \vspace{-3mm}
  \caption{On the tasks we consider, our approach approximately reproduces the network classification outputs. In particular, we obtain $>99\%$ on the sentiment analysis as it shown for the True Positive (TP) and True Negative (TN) columns vs. the False Positive (FP) and False Negative (FN) columns, where TP and TN represent the correspondence with the classification of the network over all the test set. For the ECG classification we reach $>97\%$ agreement. See App.~\ref{app:kann_quantitative}}
  \label{fig:kann_acc}
\end{figure*}

In addition to identifying principal features of beat signals, we show in what follows that the Koopman eigenvectors are also instrumental in analyzing the latent structure of the LSTM autoencoder. We begin by showing in Fig.~\ref{fig:ecg_sa} (left) the median values over time of a normal batch and its reconstruction (as in Fig.~\ref{fig:ecg_eigs}), and similarly for a batch of anomalous signals (orange and brown). From this data, the task of ECG classification may be viewed as a binary classification problem, separating normal from anomalous signals via reconstruction. However, we will now show that this is actually not the case. Instead, the network essentially encodes inputs, whether normal or anomalous, in a representation that is closer to the manifold of \emph{normal} signals. To demonstrate and analyze this phenomenon, we consider hidden state tensors $H_\text{no}$ and $H_\text{an}$ of normal and anomalous beats, respectively, and we concatenate these tensors over the samples yielding $H = (H_\text{no}, H_\text{an}) \in \mathbb{R}^{128 \times 140 \times 64}$. We would like to study the decision boundary separating between different signals in the latent space. 

To this end, we employ a standard measure known as the silhouette score~\citep{rousseeuw1987silhouettes} to quantify the class separation quality. The silhouette score $\sigma$ is a real value in $[-1, 1]$, where scores close to $1$ mean the latent states are well separated. In contrast, values closer to zero indicate that samples are located on or close to the decision boundary. We compute the silhouette score estimates on $\tilde{H} = H \cdot B$ averaged over samples, and cumulatively averaged over time. Namely, $\sigma(t) = \frac{1}{128 \cdot t} \Sigma_{s,t} \sigma(\tilde{h}_{s,t})$, where $\sigma(\tilde{h}_{s,t})$ is the silhouette score of the vector $\tilde{h}_{s,t}$ with $s \in [1,\dots,128]$ and $t\in [1,\dots,140]$. We compare three silhouette score estimates denoted by: $\sigma$ for the original $\tilde{H}$, $\sigma_\text{PCA}$ for the projection of $\tilde{H}$ onto its first five principal components, and $\sigma_\text{Koopman}$ for the projection of $\tilde{H}$ onto the first five dominant Koopman eigenvectors. The results are shown in Fig.~\ref{fig:ecg_sa} (right), where Koopman's embedding attains low scores compared to \code{PCA} and the original states. Namely, embedding the hidden states using Koopman eigenvectors reveals that the decision boundary between normal and anomalous signals is somewhat blurred, in contrast to the numerical results provided by the reconstructed signals and other embeddings. This understanding provides a rather straightforward interpretation of the model: it simply encodes the dominant components of all signals as being normal, allowing to easily identify anomalous signals later by measuring their reconstruction error. Finally, our analysis shows that Koopman eigenvectors successfully identify the salient features of normal beat signals. We conclude from this observation that the network focuses on identifying these features and reconstructing them accurately. A correct reconstruction of these salient features allows to subsequently identify using a simple loss check whether a signal is normal or anomalous. Importantly, we show in Fig.~\ref{fig:ecg_sa} that the network indeed successfully reconstructs normal and anomalous signals. The understanding that we obtain using \code{KANN} is that the network mainly focuses on these salient features during the reconstruction.

% \newpage
% \vspace{-3mm}
\subsection{\code{KANN} reproduces the latent dynamics}
\label{subsec:kann_quantitative}

We perform a quantitative study of the ability of $C$ to truly capture the latent dynamics. We show in Fig.~\ref{fig:kann_acc} that indeed, \code{KANN} is able to reproduce the nonlinear dynamics of the network in Eq.~\eqref{eq:neural_dynamics} to a high degree of precision, and thus we achieve the empirical justification to replace $F$ with $C$. Namely, we show that indeed $C$ represent the neural network at hand and not some arbitrary dynamics.

To this end, we consider the following two metrics: \textbf{Relative error} and \textbf{Accuracy error} (see App.~\ref{app:kann_quantitative}).
For the sentiment analysis problem (Fig.~\ref{fig:kann_acc}, left), we obtain $>99\%$ correspondence with the classification of the network over \emph{all} the test set.
In the ECG classification task we obtain the same result in terms of accuracy error. Further, in Fig.~\ref{fig:kann_acc}, right, we reconstruct $145$ signals of the normal test set and compute their loss.
Finally, we also computed the relative error of the hidden states, obtaining $e_\mathrm{rel}=0.095$ for the sentiment analysis task, and $e_\mathrm{rel}=0.0056$ for the ECG classification problem. For more details see App.~\ref{app:kann_quantitative}. 

% \vspace{-3mm}
\section{Discussion}

In this work we presented a novel framework for studying sequence neural models based on Koopman theory and practice. Our method involves a dimensionality reduction representation of the states, and the computation of a linear map between the current state and the next state. Key to our approach is the wealth of tools we can exploit from linear analysis and Koopman-related work. In particular, we compute linear approximations of the state paths via simple matrix-vector multiplications. Moreover, we identify dominant features of the dynamical system and their effect on inference and prediction. Our results on the sentiment analysis problem, and the ECG classification challenge provide simple yet accurate descriptions of the underlying dynamics and behavior of the recurrent models. Our work lays the foundations to further develop application-based descriptive frameworks, towards an improved understanding of neural networks. In the future, we plan to explore our framework during the training of the model, where in this work we focused only on trained models.

\clearpage

\section*{Acknowledgements}

This research was partially supported by the Lynn and William Frankel Center of the Computer Science Department, Ben-Gurion University of the Negev, an ISF grant 668/21, an ISF equipment grant, and by the Israeli Council for Higher Education (CHE) via Data Science Research Center, Ben-Gurion University of the Negev, Israel.

\bibliography{refs}

% APPENDIX
%%%%%%%%%%%%%%%%%%%%%%%%%%%%%%%%%%%%%%%%%%%%%%%%%%%%%%%%%%%%%%%%%%%%%%%%%%%%%%%
%%%%%%%%%%%%%%%%%%%%%%%%%%%%%%%%%%%%%%%%%%%%%%%%%%%%%%%%%%%%%%%%%%%%%%%%%%%%%%%

%%%%%%%%%%%%%%%%%%%%%%%%%%%%%%%%%%%%%%%%%%%%%%%%%%%%%%%%%%%%%%%%%%%%%%%%%%%%%%%
%%%%%%%%%%%%%%%%%%%%%%%%%%%%%%%%%%%%%%%%%%%%%%%%%%%%%%%%%%%%%%%%%%%%%%%%%%%%%%%

\appendix

\title{An Operator Theoretic Approach for Analyzing Sequence Neural Networks \\ Supplementary Material}
% \twocolumn[
\onecolumn

% \vskip 0.3in

\section{Unigram and bigram highlighting in sentiment analysis}
\label{app:results_sentiment_analysis}

Several examples of reviews in which the positive and negative unigrams are highlighted by the projection magnitude of the hidden states are shown in Fig.~\ref{fig:ngram_highlighting}. In particular, we note the first negative review, where the network decreases the projection magnitude when it identifies a positive word (\code{excellent}), and the magnitude increases significantly when the word \code{bad} appears. To further assess the eigenvectors role in identifying positive and negative unigrams in the reviews, we perform the following statistical evaluation. We employ a Bag Of Words (\code{BOW}) algorithm on the vocabulary to classify the sentiment on the word level. Using the \code{BOW} classification, we extract the positive and negative words from each review, and we compute how many of these words attain high projection magnitude values. Specifically, given a batch $H$, we compute the following averages
\begin{equation} \label{eq:unigram_stats}
    a_p(H) = \frac{1}{|T_p|} \sum_{t \in Tp} \lfloor s(1, h_t) \rceil \ , \quad a_n(H) = \frac{1}{|T_n|} \sum_{t \in T_n} \lfloor s(3, h_t) \rceil \ ,
\end{equation}
where $s(j, h_t) = |\hat{h}_t(j)| \in [0, 1]$ is the projection magnitude of the state $h_t$ onto the eigenvector $u_j$, $T_p$ and $T_n$ are the indices of positive and negative elements, respectively, and the operator $\lfloor \cdot \rceil$ rounds a number to the closest integer. Thus, $a_p(H)$ and $a_n(H)$ hold the average of positive and negative words whose projection magnitude is a at least $.5$ or higher. Consequently, we can view Eq.~\eqref{eq:unigram_stats} as the percentage of identified positive and negative unigrams. We show in Fig.~\ref{fig:sa_unigram_stats} histograms of $a_p(H)$ and $a_n(H)$ over the entire test set. We find that on average, $62.3\%$ and $74.0\%$ of the positive and negative words, respectively, are identified correctly by the eigenvectors. These statistics support and reinforce our observation about the role of the eigenvectors in counting positive/negative words.

\begin{figure*}[!ht]
  \centering
  \includegraphics[width=.5\linewidth]{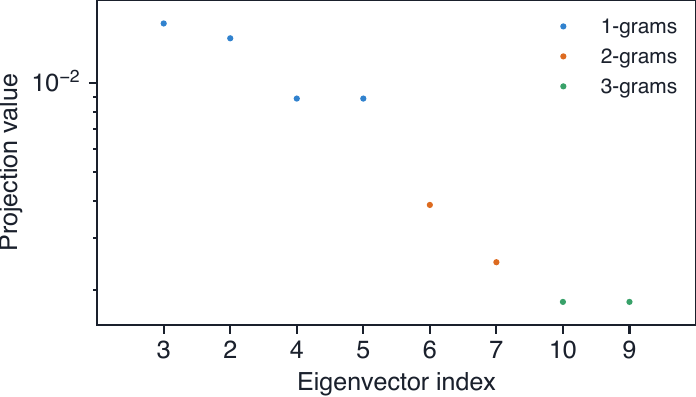}
  \caption{Projecting the batches of reviews onto the Koopman eigenvectors as specified in Eq.~\eqref{eq:eig_sort} reveals a hierarchical ordering where subspaces of eigenvectors attain different roles. In particular, the first group of eigenvectors colored blue and with indices$[3, 2, 4, 5]$ highlights $1$-grams. Similarly, the second group of eigenvectors with the orange color and indices $[6, 7]$ identifies $2$-grams. Finally, the last group in green with indices $[10, 9]$ highlights $3$-grams. We note that a similar structure was identified across different batches.}  
  \label{fig:sa_task_eig_groups}
\end{figure*}

\section{The case of general $n$-grams for $n>2$}
\label{app:results_sentiment_analysis_ngram}

In addition to unigram and bigram highlighting, we also consider \code{KANN} in the general case of $n$-grams where $n>2$. We note that $n$-grams with large $n$ are less common in sentiment analysis, both from a semantic perspective (what would a $5$-gram mean?) and also from a statistics viewpoint (e.g., even $2$-grams are scarce in the IMDB dataset). Nevertheless, we would like to show in the following that \code{KANN} indeed extends to the general case, and we test our method on $3$-gram phrases. To this end, we created a batch of $3$-grams by adding amplifiers to $2$-grams, e.g., ``not bad'' was changed to ``not extremely bad'' or ``not very bad'' and etc., and we repeated our analysis. Specifically, we extract a batch $H \in \mathbb{R}^{k \times n \times m}$ and obtain an operator $C$. Then, we project the batch on the eigenvectors as described in Eq.(10) for each $j$ eigenvector of $C$ while we sum over time and average over batch: 
\begin{equation} \label{eq:eig_sort}
        \xi_j(H) = \frac{1}{k} \sum_{h \in H} \sum_{t \in T_h} s(j, h_t) \
\end{equation}
where each $h\in H$ is a sequence of hidden states corresponds to an example in the batch, $k$ is the size of the batch and $T_h$ are the indices of the sequence. We sort $\xi(H)$ from high to low values to better visualize the importance of the Koopman eigenvectors and their ordering. We plot the resulting graph in Fig.~\ref{fig:sa_task_eig_groups} where a hierarchical behavior is shown, and each group of eigenvectors have a different role. The first group with labels $[3, 2, 4, 5]$ with the highest projection values captures $1$-grams as discussed in the main text. The second group with labels $[6, 7]$ encodes $2$-grams, and the third group with labels $[10, 9]$ encodes $3$-grams as we show in Fig.~\ref{fig:3gram_highlighting}. Specifically, using the same method we described in the main text, we obtain highlighting of $3$-gram components in the review when projected onto eigenvector $10$.  Importantly, in our analysis on $3$-grams we did not re-train the network. Rather, we simply created a batch with $3$-gram components and analyzed the results obtained with \code{KANN}.

\begin{figure*}[!ht]
  \centering
  \begin{overpic}[width=0.8\linewidth]{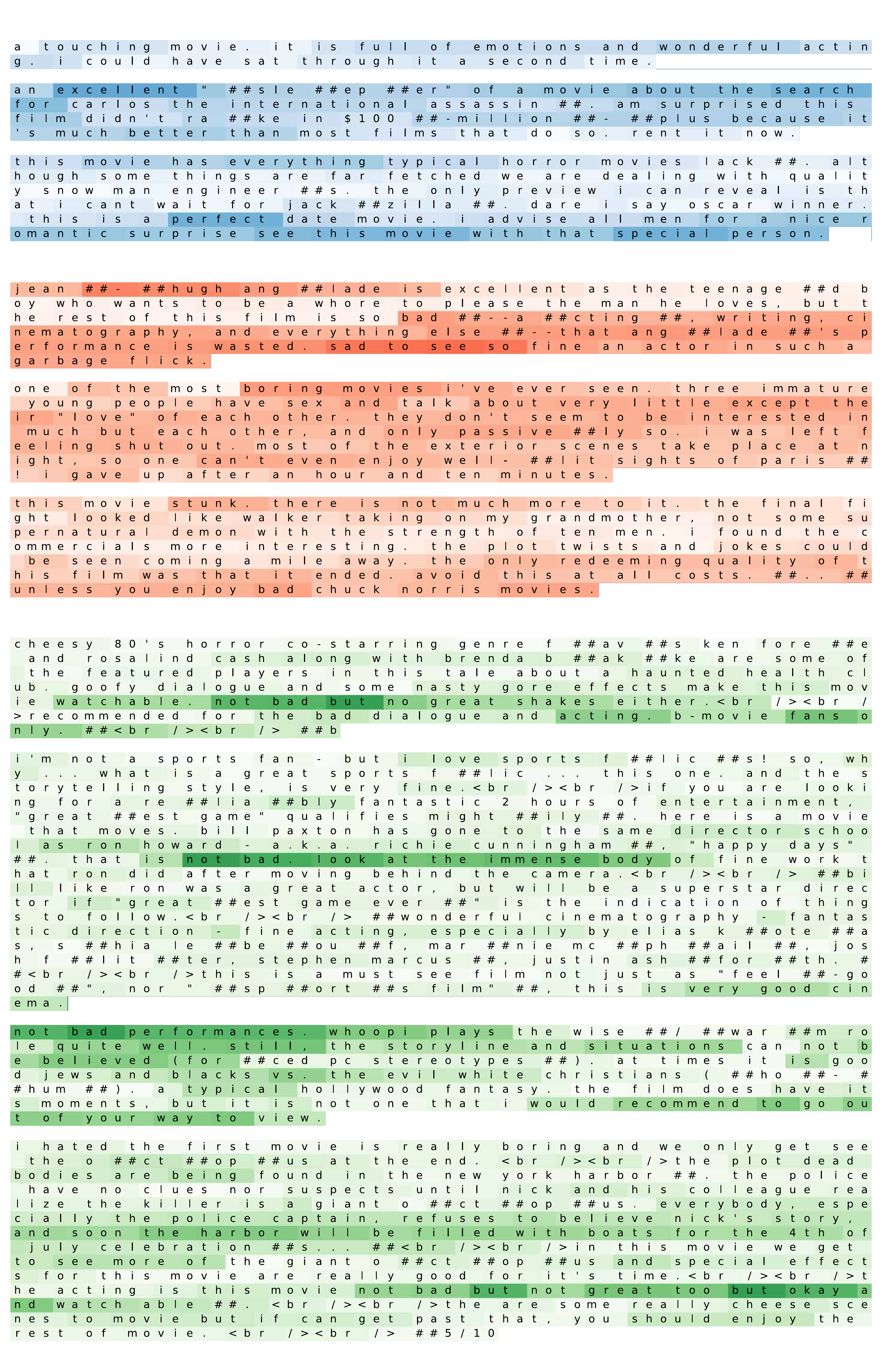} 
    \put(0, 98){Positive unigram reviews:} \put(0, 80.5){Negative unigram reviews:} \put(0, 54.5){Bigram reviews:}
  \end{overpic}
  \caption{Several examples of highlighted unigrams and bigrams.}  
  \label{fig:ngram_highlighting}
\end{figure*}

\begin{figure*}[!ht]
  \centering
  \includegraphics[width=1\linewidth]{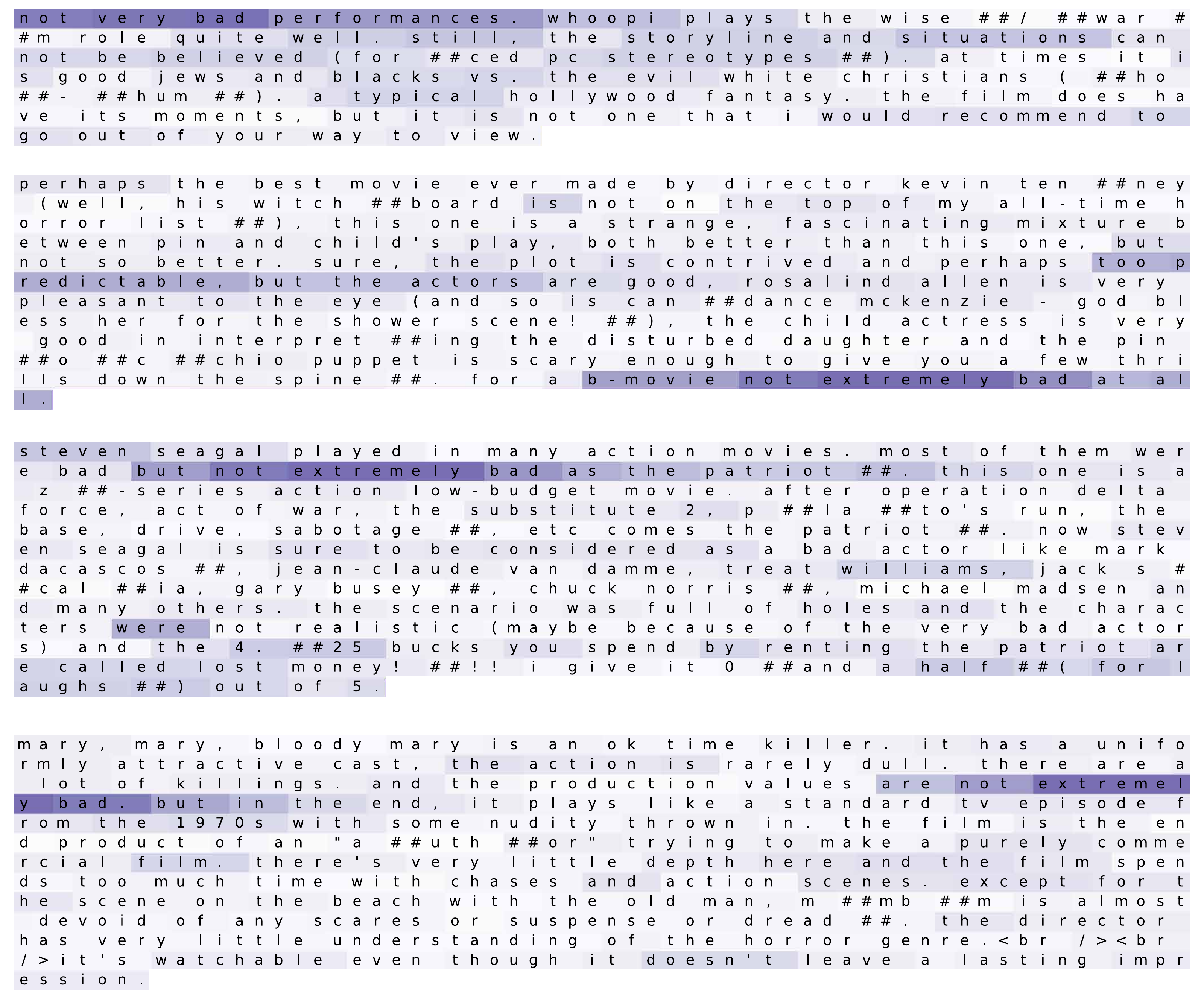}
  \caption{Projecting the hidden states to eigenvector $10$ highlights 3-gram components, similarly to the highlighting of 1-grams and 2-grams in Fig. ~\ref{fig:ngram_highlighting}.}
  \label{fig:3gram_highlighting}
\end{figure*}

\clearpage
\section{Shuffled reviews}
\label{app:sa_shuffled_revs}

We will consider the following two reviews: ``not bad and very good'' and ``very bad and not good''. These reviews serve as a great example to understand the behavior of the network as they contain the same words in a different order, and the overall meaning is opposite. Considered from a linear dynamical systems viewpoint, these reviews seem to pose a challenge: given that the tokens are the same but in a different order, how will a linear system be able to distinguish between them? Indeed, assuming a linear system in the space of tokens would be problematic in this case. However, we would like to emphasize that our main claim in the paper is that the dynamics in the \emph{latent space} are sufficiently linear to allow for Koopman analysis. As the network is recurrent and nonlinear, and the tokens are processed one-by-one, the above reviews can be distinguished in practice while not breaking the linearity in the latent space. Specifically, each of these reviews is embedded completely differently. The first example starts with the word ``not'', whereas the second example starts with the word ``very''. In practice, the network embeds these latent states in completely different locations of the latent space. Moreover, the embedding of the other states are related to the initial locations. Thus, in practice, the network has two completely different trajectories for the example reviews in discussion. From a (linear) dynamical systems perspective, we have two trajectories of different initial conditions. Such cases can be typically differentiated and identified using linear dynamical systems. 

To show it numerically, we generated these two reviews (five words each), and we repeated our analysis. Specifically, we projected their latent trajectories (as obtained from the network) onto the first two dominant \code{PCA} modes (as was done in~\cite{maheswaranathan2019reverse}), and we report the obtained paths in Tab.~\ref{tab:sa_paths}. Keeping in mind that the network is mostly one-dimensional in a \code{PCA} representation, the above paths clearly show that the reviews are correctly classified. Namely, the first review starts in the negative part of the $x$-axis ($-0.716$) and finishes in the positive part of the $x$-axis ($0.713$). In comparison, the second review starts in the positive part of the $x$-axis ($1.478$) and finishes in the negative part of the $x$-axis ($-0.572$). Computing the paths using our \code{KANN} representation via the matrix $C$, we obtain the paths reported in Tab.~\ref{tab:sa_paths}. While the values are different (as our linear approximation exhibits some error), the initial positions and trend are the same for the nonlinear representation and our Koopman linear representation. In particular, the trajectories end on the same side of the x-axis for each review, exactly as we have in the nonlinear network.

\begin{table*}[ht]
\centering
\begin{tabular*}{\textwidth}{@{\extracolsep{\stretch{1}}}*{1}{l}*{4}{cc}@{}}
    \toprule
            & \multicolumn{2}{c}{``not bad and very good''} & \multicolumn{2}{c}{``very bad and not good''} \\ \cmidrule(lr){2-3}\cmidrule(lr){4-5}
    Time    & \code{Network} & \code{KANN} & \code{Network} & \code{KANN} \\
    \midrule
    $t=1$   & $(-0.716, \; 0.547)$      & $(-0.284, \; 0.649)$  & $(1.478, \; 0.355)$   & $(1.094, \; 0.522)$ \\
    $t=2$   & $(-1.080, \; 0.280)$      & $(-0.589, \; 0.612)$  & $(0.313, \; 0.078)$   & $(0.128, \; 0.619)$ \\
    $t=3$   & $(-0.440), \; -0.093)$    & $(0.003, \; 0.479)$   & $(0.392, \; -0.282)$  & $(0.335, \; 0.503)$ \\
    $t=4$   & $(0.807, \; -0.319)$      & $(0.708, \; 0.419)$   & $(-0.895, \; -0.311)$ & $(-0.530, \; 0.571)$ \\
    $t=5$   & $(0.713, \; 0.036)$       & $(0.598, \; 0.474)$   & $(-0.572, \; -0.291)$ & $(-0.237, \; 0.542)$ \\
    \bottomrule
\end{tabular*}

\caption{The nonlinear network as well as our linear representation are able to differentiate between the reviews ``not bad and very good'' and ``very bad and not good'', and to correctly classify them. Specifically, we show the trajectories of the hidden states as obtained from the network and our method when projected to the first dominant \code{PCA} modes. The results above show similar initial conditions and trend, i.e., both start and end on the same side of the $x$-axis. We conclude that the network learns a representation which is sufficiently linear in the latent space, allowing to methods such as ours to expose its dynamics.}
\label{tab:sa_paths}
\end{table*}

% \clearpage
% \vspace{-3mm}
\section{Projecting normal beat signals onto \code{PCA} components}
\label{app:results_ecg}

\begin{figure*}[ht]
  \centering
  \includegraphics[width=1\linewidth]{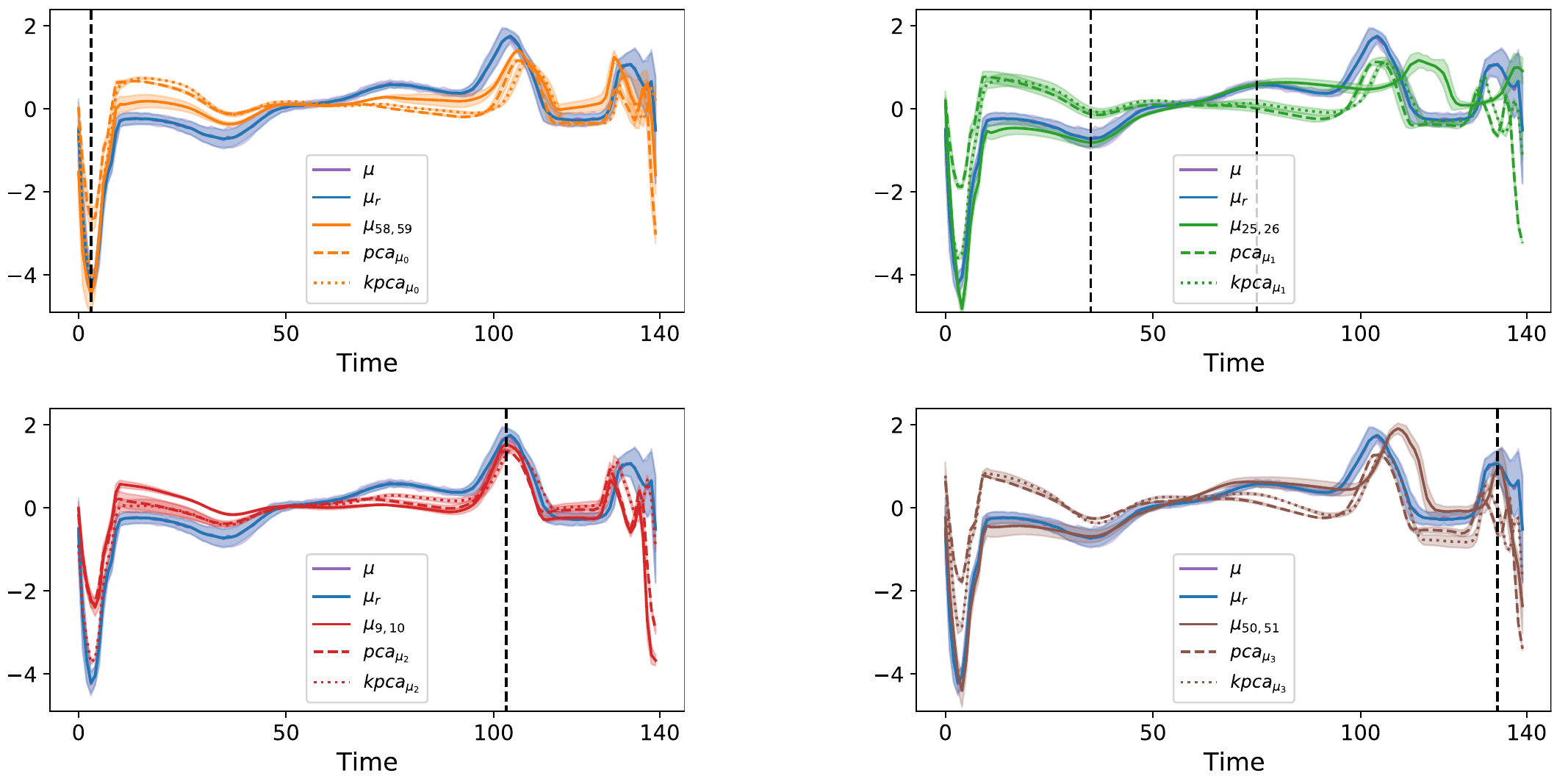}
  \caption{We show the first four principal modes of \code{KANN} (solid lines), \code{PCA} (dashed lines), and \code{KernelPCA} (dotted lines). The above graphs show that our method is better at matching the salient features of beat signals which are marked by black dashed lines in comparison to \code{PCA} and \code{KernelPCA}. We conclude that the network mainly focuses on reconstructing these salient features, allowing the user to easliy distinguish between normal and anomalous beats during post-prcoessing.} 
  \label{fig:ecg_eig_cmp}
\end{figure*}

In Sec. $4.2$ in the main text, we discover that the dominant Koopman eigenvectors are capable of identifying the salient features in the beat signals (marked by dashed black lines in Fig.~\ref{fig:ecg_eig_cmp}). To compare our results with \code{PCA} and \code{KernelPCA} (using \code{rbf} kernel), we now repeat the same experiment, but instead of projecting onto Koopman modes, we project the hidden states $H$ to the first four PCs and first four eigenvectors of the centered kernel matrix respectively. We provide both qualitative and quantitative comparison with both methods. Fig.~\ref{fig:ecg_eig_cmp} shows the resulting graphs, clearly demonstrating that \code{PCA} and \code{KernelPCA} fail to encode the dynamics. In Tab.~\ref{tab:ecg_comp} we provide a quantitative comparison of our method to \code{PCA} and \code{KernelPCA}. Specifically, for every method, we compute the mode with the minimal distance to the salient features located at times $t=3, \; 35\leq t \leq 75, \; t=103$ and $t=133$. The results clearly show that \code{KANN} attains the lowest error for each of the salient features.

\begin{table*}[!ht]
\centering
\begin{tabular*}{\textwidth}{@{\extracolsep{\stretch{1}}}*{1}{l}*{2}{cc}@{}}
    \toprule
    Method & $t=3$ & $ 35 \leq t \leq 75$ & $t=103$ & $t=133$ \\
    % \midrule
    % \code{PCA} & $1.5817$ & $1.9728$ & $0.3356$ & $1.6891$ \\
    % \code{KernelPCA} & $0.0762$ & $2.0850$ & $0.4095$ & $1.0329$ \\
    % \textbf{\code{KANN}} & $0.2316$ & $\boldsymbol{0.5107}$ & $\boldsymbol{0.1724}$ & $\boldsymbol{0.0871}$ \\
    \midrule
    \code{PCA} & $1.5817$ & $1.2197$ & $0.3356$ & $1.1685$ \\
    \code{KernelPCA} & $0.0762$ & $1.1045$ & $0.4095$ & $0.7217$ \\
    \textbf{\code{KANN}} & $\boldsymbol{0.0317}$ & $\boldsymbol{0.5107}$ & $\boldsymbol{0.1724}$ & $\boldsymbol{0.0871}$ \\
    \bottomrule
\end{tabular*}

\caption{For every salient feature at times $t=3, \; 35\leq t \leq 75, \; t=103$ and $t=133$, we compute the distance between the signal and its reconstruction using the principal modes of \code{PCA}, \code{KernelPCA} and \code{KANN}. Our approach exhibits the minimal error in comparison to \code{PCA} and \code{KernelPCA}.}
\label{tab:ecg_comp}
\end{table*}

% \begin{figure*}[ht]
%   \centering
%   \includegraphics[width=1\linewidth]{figures/ecg_pca}
%   \caption{Unlike Koopman's eigenvectors, projecting normal beats on the principal components does not reveal any particular information that is related to the reconstructed signals.}  
%   \label{fig:ecg_pca}
% \end{figure*}

\begin{figure*}[ht]
  \centering
  \includegraphics[width=.5\linewidth]{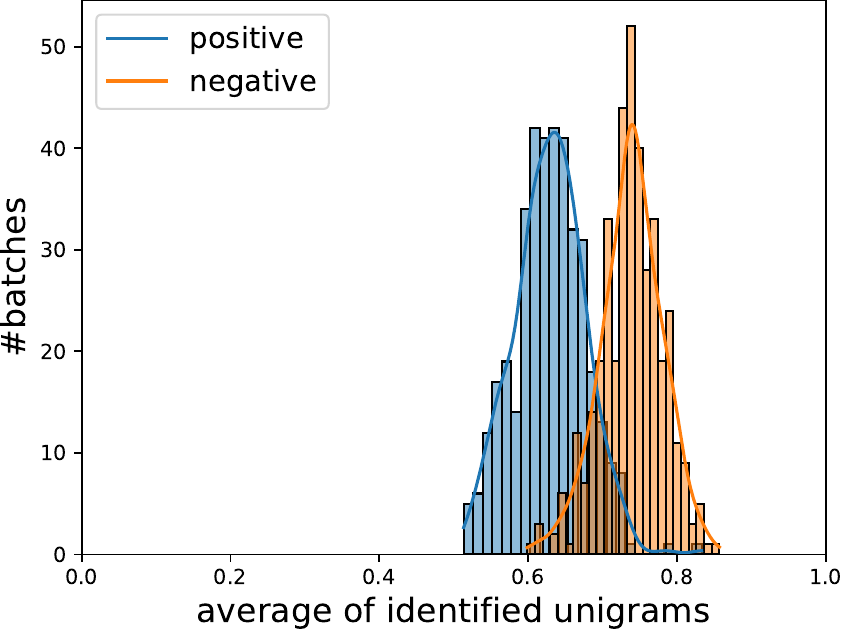}
  \caption{The above histograms show the average percentage of identified positive (blue) and negative (orange) unigrams per batch. It follows that negative words are better identified by the network ($74.0\%$) in comparison to positive words ($62.3\%$).}  
  \label{fig:sa_unigram_stats}
\end{figure*}

\clearpage

\section{Different basis and network architectures}
\label{app:basis_and_archs}

\begin{figure*}[ht]
  \centering
  \begin{overpic}[width=.85\linewidth]{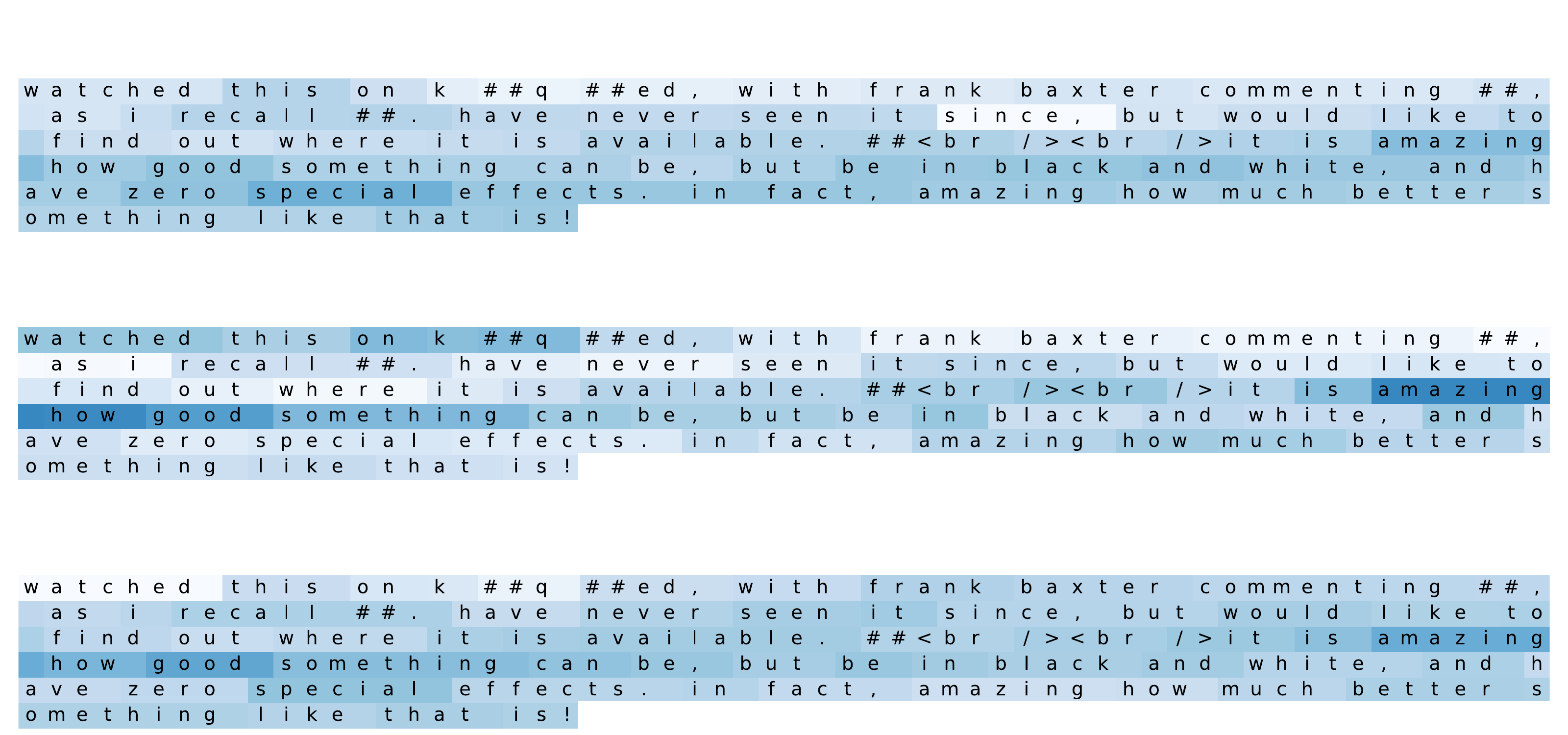} 
    \put(0, 44){Using \code{SVD}:} \put(0, 28){Using \code{PCA}:} \put(0, 12){Using \code{FFT}:}
  \end{overpic}
  \caption{Computing $C$ using \code{SVD}, \code{PCA}, and \code{FFT} yield eigenvectors with the same semantic role. Indeed, projections using different bases highlight various positive words in the same review.}  
  \label{fig:sa_cmp_basis}
\end{figure*}

\begin{figure*}[ht]
  \centering
  \includegraphics[width=1\linewidth]{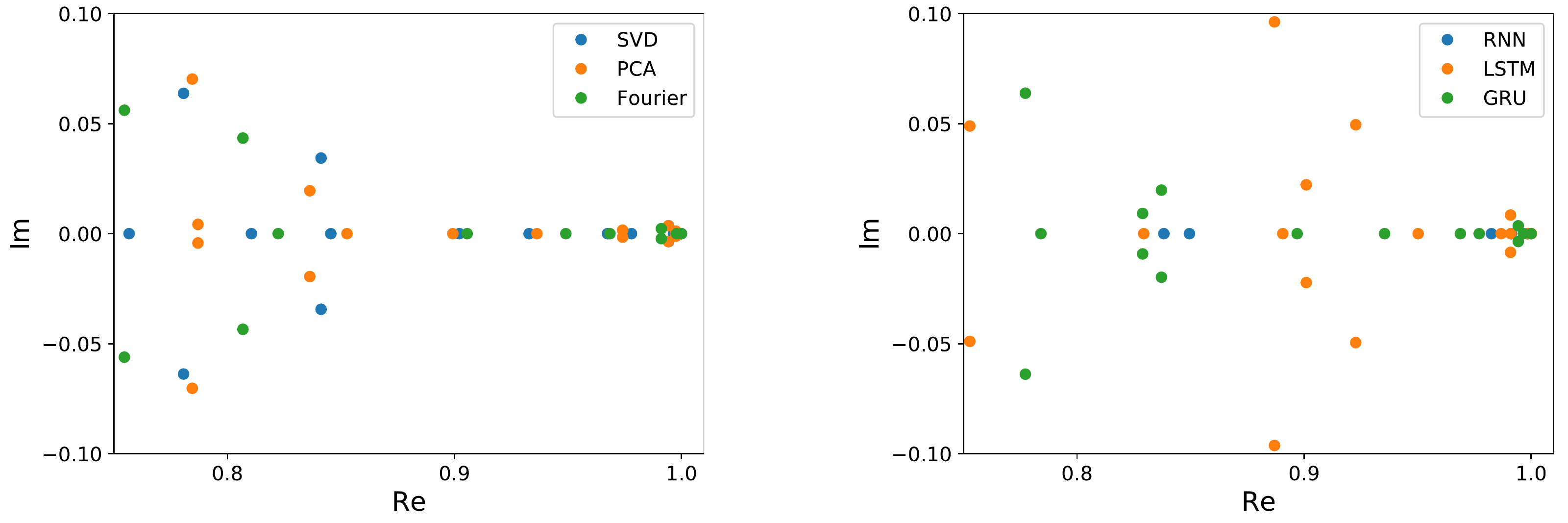} 
  \caption{We show the dominant eigenvalues of various $C$ matrices in the complex plane ($\mathrm{Re}$, $\mathrm{Im}$). Most of the eigenvalues correspond when different bases such as \code{SVD}, \code{PCA}, \code{FFT} are used (left). Similarly, using recurrent components such as \code{RNN}, \code{LSTM}, \code{GRU} leads to related spectra (right).}  
  \label{fig:sa_cmp_eig}
\end{figure*}

\paragraph{Choice of basis.} We will now demonstrate the robustness of our approach to the choice of basis. The first step to computing the matrix $C$ involves the projection of the given states onto a basis. In our work, we mostly experimented with the truncated \code{SVD} modes obtained by decomposing the hidden states tensor. In what follows, we additionally show that the principal component analysis (\code{PCA}), and Fourier transform (\code{FFT}) bases lead to quantitatively similar results on the sentiment analysis task. We note that while the bases are linear in terms of projection, \code{SVD} and \code{PCA} are data-driven, whereas \code{FFT} is data-agnostic, i.e., the basis elements are independent of the data. First, we compute the relative error as in Sec. 4.4, and we obtain $\num{0.0347}, \num{0.0347}, \num{0.973}$ for \code{SVD}, \code{PCA}, and \code{FFT}, respectively. The somewhat poor result of \code{FFT} is expected, as it is data-agnostic. Second, we compare the dominant eigenvalues of the different $C$ matrices computed using the bases. It follows that across all bases, the dominant eigenvalues correspond to one another. In particular, the average error between corresponding eigenvalues is $0.003$ for \code{PCA}, and $0.02$ for \code{FFT}, when measured from the eigenvalues of \code{SVD}. We additionally plot the dominant eigenvalues in Fig.~\ref{fig:sa_cmp_eig} where the $x$-axis is the real part, and the $y$-axis is the imaginary part. Finally, we also show how the dominant eigenvectors have the same semantic role in highlighting the positive words in the same review. Indeed, we show in Fig.~\ref{fig:sa_cmp_basis} that the positive words obtain large projection magnitudes in all bases. See the words e.g., \code{amazing}, \code{special}, \code{good}. Overall, the results show robustness to linear bases.

% \vspace{-5mm}

% \clearpage

\begin{figure*}[t]
  \centering
  \begin{overpic}[width=1\linewidth]{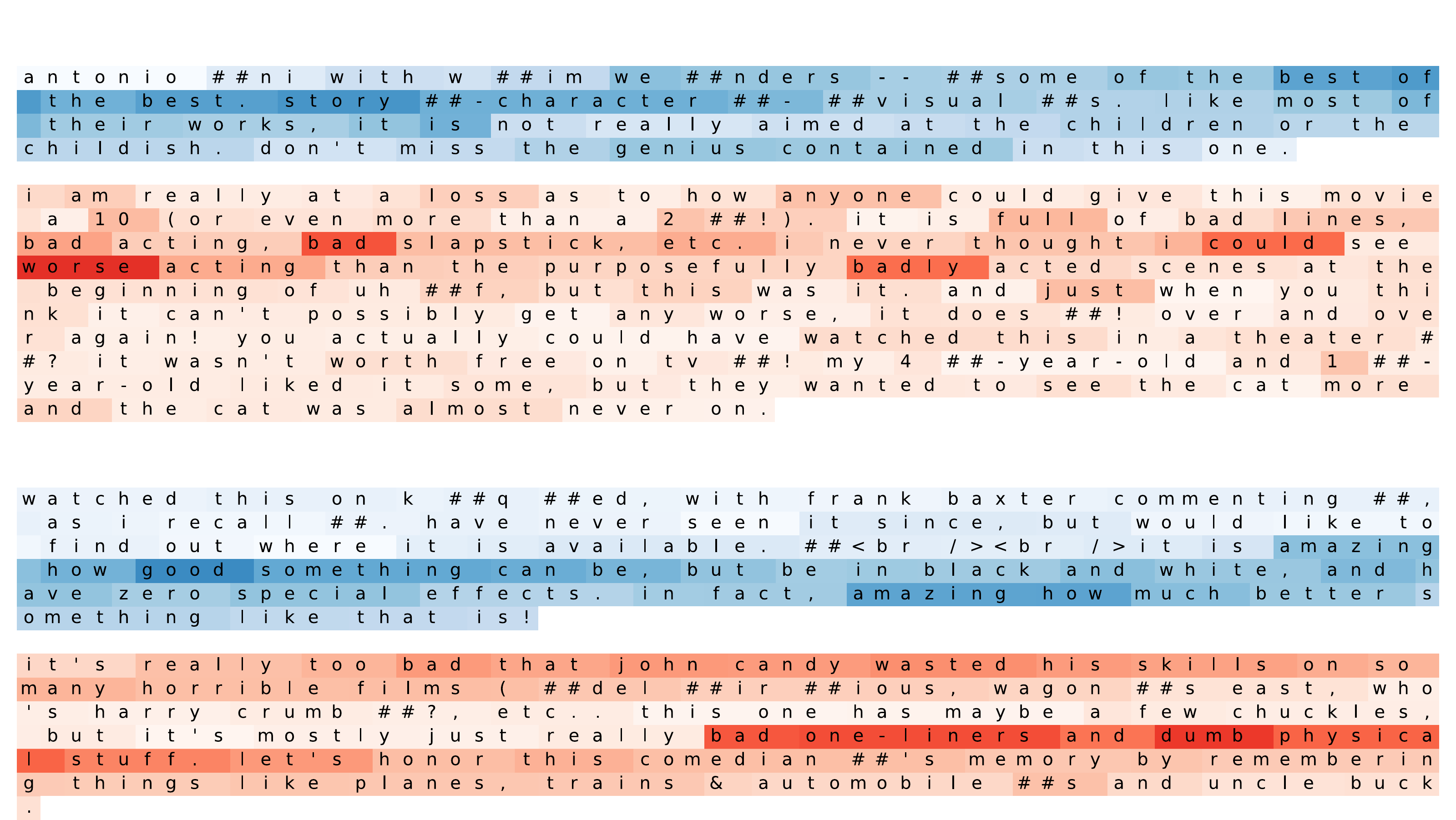} 
    \put(0, 53){\code{RNN} model:} \put(0, 24){\code{LSTM} model:}
  \end{overpic}
  \caption{Examples of highlighted unigrams obtained from the trained \code{RNN} and \code{LSTM} models.}  
  \label{fig:unigram_archs}
\end{figure*}

\paragraph{Results extend across architectures.} In addition to robustness to the basis, we also verify that our results qualitatively extend across different architectures. Specifically, we trained a vanilla recurrent neural network (\code{RNN})~\citep{elman1990finding}, a long short term memory model (\code{LSTM})~\citep{hochreiter1997long}, and a gated recurrent unit (\code{GRU}) network on the sentiment analysis problem. Then, we extract a single batch from the test set, and evaluate our \code{KANN} approach on the trained models. We find that our analysis yields similar results in all cases. In particular, the dominant eigenvalues of each of the models attain related values as can be seen in Fig.~\ref{fig:sa_cmp_eig} (right). Moreover, we find that the dominant eigenvectors share the same role of highlighting positive and negative unigrams. To verify this, we computed the histograms of identified positive and negative words as in Fig.~\ref{fig:sa_unigram_stats}. We observe that on average $62\%$, $76\%$, and $62\%$ positive words are discovered by the projection magnitude of the \code{RNN}, \code{LSTM}, and \code{GRU} models. Similarly, the negative unigrams are highlighted in an average of $55\%$, $83\%$, and $74\%$ for \code{RNN}, \code{LSTM}, and \code{GRU}. Indeed, there is a large variation in the statistics of the models, where \code{LSTM} obtains the best averages, followed by \code{GRU}, and \code{RNN} is last. Nevertheless, in all cases, the average identification of unigrams is above $50\%$, and given that \code{BOW} is noisy by itself, we believe these statistics are qualitatively similar. In addition, we plot in Fig.~\ref{fig:unigram_archs} a few examples of highlighted reviews obtained with the models.

% explain the dynamics of the copy problem: measure-preserving map, 
\section{Results on the copy task}
\label{app:copy_task}

The copy task was designed to test the memory retaining capabilities of recurrent units~\citep{hochreiter1997long}. In this task, the network is expected to memorize the first few characters in the input array and copy them to the end of the output vector which is otherwise filled with blanks. For instance, the input-output structure reads $\texttt{928-{}-{}-{}:-{}-} \quad \mapsto_\varphi \quad \texttt{-{}-{}-{}-{}-{}-928}$, if the model is required to remember three digits across three blanks. Thus, the challenge increases with more digits to remember and when the amount of blanks is higher. We trained a \code{dtriv} architecture~\citep{casado19trivializations} on the copy task with three characters to remember and $30$ blanks for $500$ iterations. The \code{dtriv} model is similar to a vanilla RNN with the exception that its hidden-to-hidden transformation is \emph{orthogonal}. The network converges to an accuracy of $100\%$ on the training and test data as it is a relatively easy setting. The following analysis is based on a test batch of size $32$, yielding a states tensor $H \in \mathbb{R}^{32 \times 36 \times 48}$ where the middle dimension is the sequence length, and the last dimension is the hidden state size. 

\paragraph{The latent structure of the copy task is measure-preserving.} Our first analysis result deals with the geometric structure of the learnt dynamics. Before discussing our results, we make the following three observations. First, the copy problem with its unknown dynamics $\varphi$ which maps inputs to outputs, is isometric. Indeed, for many choices of norms, e.g., $L^2$, it follows that $d(x_1, x_2) = d(y_1=\varphi(x_1), y_2=\varphi(x_2))$ where $x_1, x_2$ are two input vectors, and $y_1, y_2$ are two output vectors. Thus, $\varphi$ belongs to the class of \emph{measure-preserving} dynamical systems. Second, while \code{dtriv} uses orthogonal hidden-to-hidden matrices, the overall network transformation is not necessarily isometric due to the nonlinear activation layers. Indeed, the analysis in~\citep{arjovsky2016unitary} which also applies to \code{dtriv} only establishes an upper bound on the gradient norms, and there is no lower bound. In practice, since \code{dtriv} uses \code{modReLU}~\citep{arjovsky2016unitary}, it may cause a non-isometric transformation to the latent space. Third, Koopman theory establishes a connection between the algebraic properties of the linear operator to the geometric structure of the dynamics as we show next.  The following result is not novel~\citep{eisner2015operator}, but we prove it below for completeness.
\begin{prop} \label{prop:unitary}
Let $\varphi$ be an invertible measure preserving dynamical system on a compact, inner-product domain $\mathcal{M}$. Then its associated Koopman operator is unitary.
\end{prop}

\paragraph{Proof.} 

Let $\varphi: \mathcal{M} \rightarrow \mathcal{M}$ be a map on the compact, inner-product space $\mathcal{M}$. We denote by $\mu$ the continuous measure on $\mathcal{M}$, and its induced metric $\| z \|$. The map $\varphi$ is measure preserving, i.e., $\mu(\varphi^{-1} A) = \mu(A)$ for every measurable set $A \subset \mathcal{M}$. Let $\mathcal{K}_\varphi$ be the Koopman operator of $\varphi$ acting on the function space of square integrable function $L^2$. Given the indicator function $1_A$ for the set $A$, we have that
\[
    \mathcal{K}_\varphi 1_A (z) = 1_A(\varphi \circ z) = 1_{\varphi^{-1} A} (z) \ ,
\]
and thus 
\begin{align*}
    \int_\mathcal{M} \mathcal{K}_\varphi 1_A \dd \mu &= \mu(\varphi^{-1} A) = \mu(A) = \int_\mathcal{M} 1_A \dd \mu \ .
\end{align*}
Moreover, positive functions converge to a representation using simple indicator functions. Consequently, we have that $\int_\mathcal{M} \mathcal{K}_\varphi f \dd \mu = \int_\mathcal{M} f \dd \mu$ for general $f \in L^2$ since it can be written as the difference of the integrable negative and positive components of $f$.

The Koopman operator is linear and it is pointwise multiplicative, i.e., $\mathcal{K}_\varphi (\alpha f + \beta g) = \alpha \mathcal{K}_\varphi(f) + \beta \mathcal{K}_\varphi(g)$ and $\mathcal{K}_\varphi (f\,g) = \mathcal{K}_\varphi (f) \mathcal{K}_\varphi(g)$, where $\alpha, \beta \in \mathbb{R}$, and $f,g \in L^2$. Due to these observations, it follows that $\mathcal{K}_\varphi$ preserves the inner product of functions, namely, for every $f, g \in L^2$
\begin{align*}
    \langle f, g\rangle = \int_\mathcal{M} f \, g \dd \mu = \int_\mathcal{M} \mathcal{K}_\varphi (f \, g) \dd \mu = \langle \mathcal{K}_\varphi(f),\, \mathcal{K}_\varphi(g) \rangle \ .
\end{align*}
Thus, the Koopman operator in this case is an isometry, since 
\begin{align*}
    \dd(f,\, g) &= \| f-g \| = \langle f-g,\, f-g\rangle^{\frac{1}{2}} = \langle \mathcal{K}_\varphi(f-g),\, \mathcal{K}_\varphi(f-g) \rangle^{\frac{1}{2}} \ .
\end{align*}
Finally, if $\varphi$ is invertible then $\mathcal{K}_{\varphi}^* \mathcal{K}_\varphi = \mathcal{K}_\varphi \mathcal{K}_{\varphi}^*$ where $\mathcal{K}_\varphi^*$ is the adjoint operator, and thus $\mathcal{K}_\varphi$ is unitary.

Given $H$ as specified above and its corresponding $C$, we find that $C$ is approximately orthogonal, i.e., $C^T C \approx \mathrm{id}$. Specifically, the relative error $|C^T C - \mathrm{id}|^2 / | C|^2 = 0.0625$. Our findings align with prior work~\citep{rustamov2013map} which shows that approximate Koopman operators are approximately orthogonal for measure-preserving maps. Therefore, although \code{dtriv} is not guaranteed to learn a measure-preserving latent map, it does so in practice as our method reveals.

\begin{wrapfigure}{r}{0.3\textwidth}
  \begin{center}
    \begin{overpic}[width=0.25\textwidth]{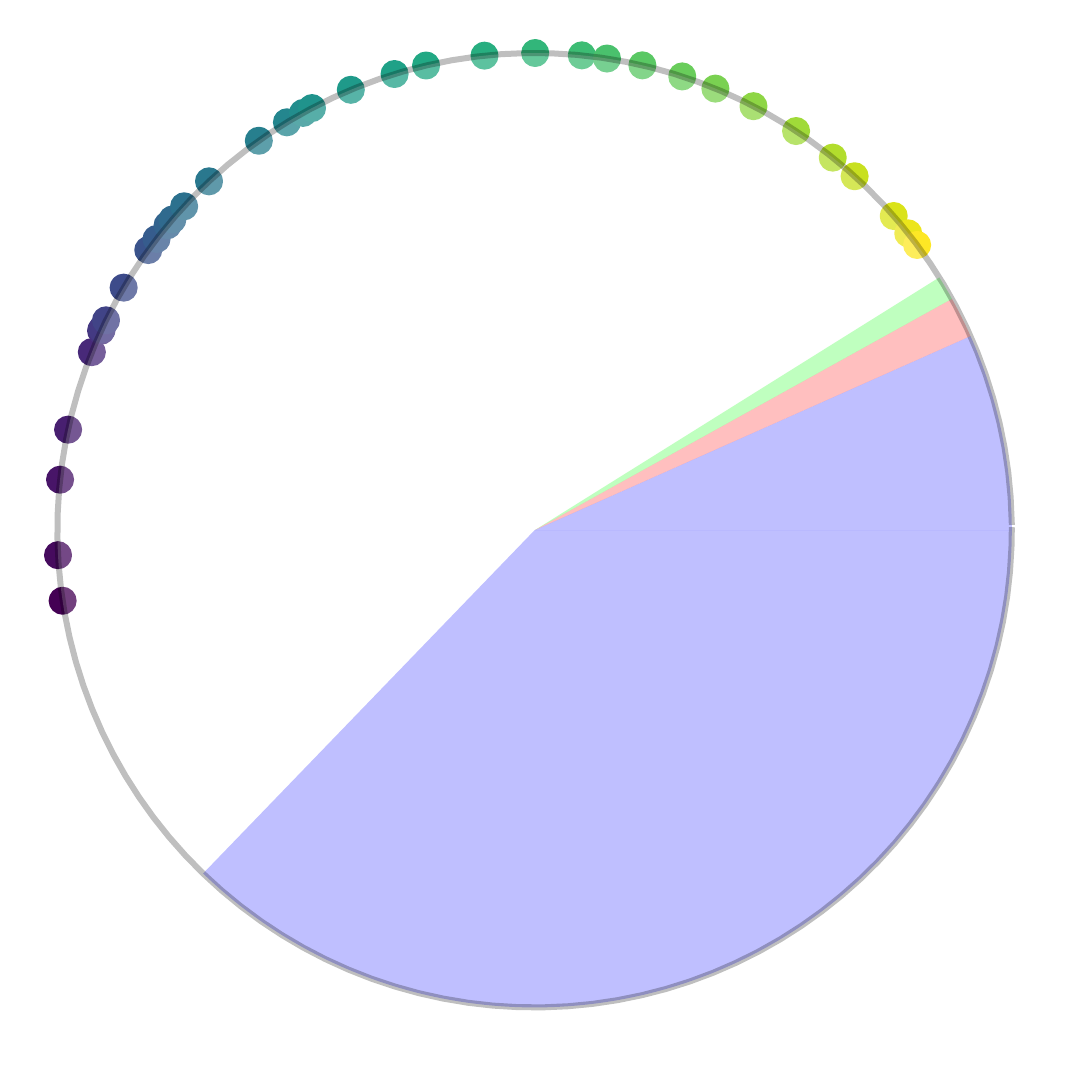}
        \put(35, 70){$0$} \put(65, 35){$2$} \put(91, 70){\tiny $3$} \put(80, 70){\tiny $5$} \put(-8, 39){$\hat{h}_0$} \put(30, 98){$\hat{h}_t$} \put(85, 81){$\hat{h}_T$}
    \end{overpic}
  \end{center}
\end{wrapfigure}
\begin{figure*}[ht]
  \centering
  \begin{overpic}[width=1\linewidth]{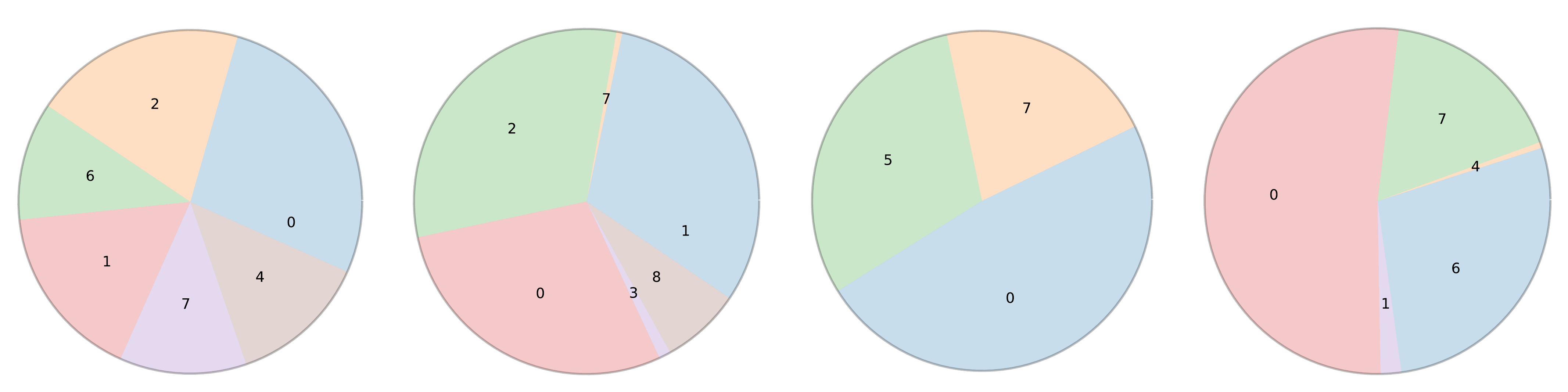} 
    \put(0, 22){$u_2$} \put(26, 22){$u_4$} \put(52, 22){$u_6$} \put(76, 22){$u_8$}
  \end{overpic}
  \caption{Every eigenvector in the copy task span multiple characters in the alphabet, allowing it contribute to the propagation of the initial digits over the sequence.}  
  \label{fig:copy_eigs_digits}
\end{figure*}

\paragraph{Eigenvectors span multiple digits in the copy task.} Our second analysis result on the copy problem focuses on the eigendecomposition of $C$. We find that most of the eigenvalues, $44$ out of $47$, are approximately unit length, i.e., $|\lambda_j -1 | < \num{5e-2}$, which also reinforces the above findings. Based on Eq.~$(9)$ it follows that the eigenvectors of those eigenvalues have long memory horizons, e.g., $\tau= 418$ for $\epsilon= \num{1e-1}$. This is well beyond the required memory horizon for this task which is $30$ as the number of blanks. Additionally, we find that all eigenvectors have the capacity to represent several characters, depending on the root of unity they are multiplied with. Namely, computing the output of the state $\tilde{h} = \mathrm{Re}(z v_j)$ for several $z$ values, yields various digits. For instance, the inset shows a specific eigenvector and its associated digits with their respective span of the unit ball. We also plot in shaded dots the coefficients of a particular input over time. Evidently, the shown eigenvector is responsible to output the blank part of the output since the coefficients are located in the zero regime. Qualitatively similar results were obtained for the other eigenvectors as we show in Fig.~\ref{fig:copy_eigs_digits} the sets of digits for select eigenvectors. For instance, $v_6$ spans the digits $\{0, 5, 7\}$, depending on the root of unity $z_i$ we multiply with $v_6$. Thus, the network essentially splits the latent space onto digit regions. Then, given an input such as $\texttt{928-{}-{}-{}:-{}-}$, the network generates its latent trajectory by carefully scaling the eigenvectors to point to the required output for every time sample.

\begin{figure*}[ht]
  \centering
  \begin{overpic}[width=1\linewidth]{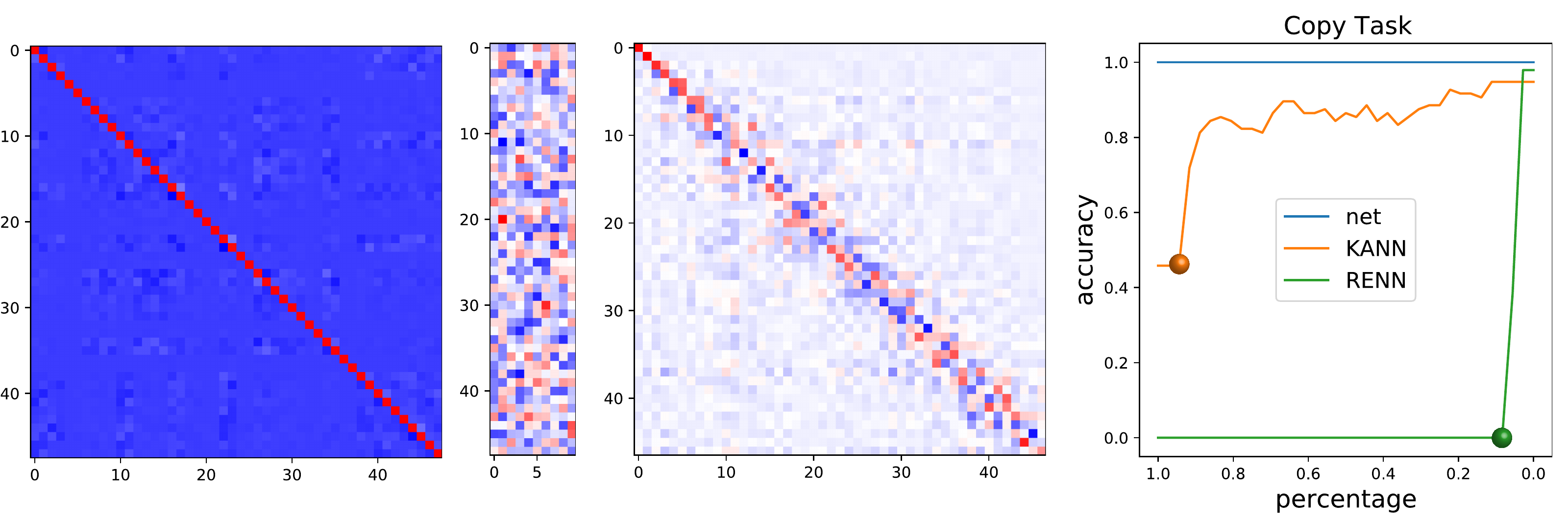} 
  \put(13,32){$\mathcal{J}^{\mathrm{rec}}$} \put(32,32){$\mathcal{J}^{\mathrm{inp}}$} \put(52,32){$C$}
  \end{overpic}
  \caption{Computing \code{RENN} components for the copy task leads to an almost identity recurrent Jacobian, $|\mathcal{J}^{\mathrm{rec}} - \mathrm{id}| = 0.11$ relative error. In comparison, our matrix $C$ is approximately orthogonal and it exhibits a diagonally-dominant structure. Our \code{KANN} approach attains good accuracy results when used to predict the states path. See the text.}  
  \label{fig:copy_task_ops}
\end{figure*}

\paragraph{Quantitative results on the copy task.} We briefly recall that \code{RENN} uses the hidden state tensor $H$ to generate a set of fixed points, i.e., points $h^*$ for which the dynamical system $h_t = F(h_{t-1}, x_t)$ is stationary $h^* \approx F(h^*, 0)$~\citep{sussillo2013opening}. Then, they derive their analysis using the input and recurrent Jacobians of $F$, $\mathcal{J}^{\mathrm{inp}}$ and $\mathcal{J}^{\mathrm{rec}}$, evaluated at a single point $(h^*,\, x^* \equiv 0)$. We show in Fig.~\ref{fig:copy_task_ops} the resulting Jacobian matrices using \code{RENN} where $\mathcal{J}^{\mathrm{rec}} \approx \mathrm{id}$ matrix (left). This is actually the expected result---as the blanks are mapped to zeros in this task, using $x^* \equiv 0$ means we look for fixed points $h^*$ related to a blank input. However, the output for a blank input should be blank as well, and thus the hidden states converge to a section of the manifold which is indifferent to the inputs. Indeed, in~\citep{sussillo2013opening,  maheswaranathan2020recurrent, maheswaranathan2020reverse}, the authors discuss approaches to select input dependent initial points $x^*$, however, it remains unclear how to avoid the above issue since any chosen point is related to a particular potential input. For reference and comparison, we show in Fig.~\ref{fig:copy_task_ops} (middle) the algebraic structure of our $C$ matrix.

To assess the information encoded in $\mathcal{J}^{\mathrm{rec}}$ and $\mathcal{J}^\mathrm{inp}$ vs. $C$, we perform the following experiment. Let $\{ h_t \}$ denote the nonlinear path of hidden states obtained from the copy task network. Given a certain threshold $l=1,...,T$, we split the path to two segments $\{ h_t \}_{t=1}^l$ and $\{ B \, C^k \tilde{h}_l \}_{k=1}^{T-l}$. That is, the first segment is simply the original states, and the second segment includes linear predictions with $C^k$ while always using $h_l$. We denote by $H^\code{KANN}_l$ the union of the paths, i.e.,
\begin{equation}
    H^\code{KANN}_l = \{ h_1, ..., h_l, B \, C \tilde{h}_l, .., B \, C^{T-l} \tilde{h}_l \} \ .
\end{equation}
For every admissible $l$, we generate $H_l^\code{KANN}$, and we compute the accuracy obtained by the network using the path $H_l^\code{KANN}$. Fig.~\ref{fig:copy_task_ops} (right) shows in blue the accuracy for the nonlinear path which is simply $100\%$. We show in orange the accuracy obtained for several $l/T$ values. The accuracy results of \code{KANN} are extremely good, even when the percentage is high, i.e., most of the path does \emph{not} use the states provided by the network, but rather, their linear prediction. Further, we emphasize that the orange point marks the percentage for three hidden states. That is, our method gets more than $80\%$ accuracy exactly when all the non-blank input digits are implicitly available in the states. Therefore, our results highlight that $C$ truly mimics the nonlinear dynamics as it is the minimal set of necessary inputs for a meaningful prediction.

In comparison, \code{RENN} can be used in a similar fashion to generate $H_l^\code{RENN}$ using the following formula 
\begin{align} \label{eq:renn_linear_state}
    h_{t+1}^{\code{RENN}} &:= h^* + \mathcal{J}^{\mathrm{rec}}(\bar{h}_t - h^*) + \mathcal{J}^{\mathrm{inp}} x_t \\
    &\approx \bar{h}_t + \mathcal{J}^{\mathrm{inp}} x_t \ ,
\end{align}
where $\bar{h}_t$ can be the original $h_t$ or $h_t^{\code{RENN}}$ depending on $l$, and the bottom formula is relevant when $\mathcal{J}^{\mathrm{rec}} \approx \mathrm{id}$ matrix. The green curve in Fig.~\ref{fig:copy_task_ops} shows the accuracy results of \code{RENN}. Due to the trivial nature of $\mathcal{J}^{\mathrm{rec}}$, \code{RENN} achieves zero accuracy in most cases, and it significantly improves when the last three states become available (marked by the green point). Thus, \code{RENN} requires almost the entire sequence of ground-truth hidden states to produce good accuracy measures in this scenario.

\section{Training Information}
\label{app:training_info}

In Tab.~\ref{tab:params} we add details regarding the models training process across each architecture and task. In the tasks column, SA, ECGC, and CT are acronyms for Sentiment Analysis, ECG Classification, and Copy Task, respectively. In addition, we used weight decay regularization in both ECG classification and Sentiment Analysis tasks.
\vspace{-4mm}
\begin{table*}[ht]
    \caption{The following hyperparameters per task and model were used during training.}
    \vspace{+0.3cm}
    \centering
    \begin{tabular}{cccccccc}
        \toprule
        Task                & Architecture  & \#epochs  & \#units   & Optimizer & LR & LR Scheduler & Clip \\
        \midrule
        SA  & \code{RNN}    & $7$       & $128$     & \code{Adam} & $\num{5e-3}$ & \code{ExpLR}, $\gamma = \num{.6}$ & $15$ \\ 
        SA  & \code{GRU}    & $5$       & $256$     & \code{Adam} & $\num{5e-3}$ & \code{ExpLR}, $\gamma = \num{.5}$ & $15$ \\ 
        SA  & \code{LSTM}   & $5$       & $256$     & \code{Adam} & $\num{1e-3}$ & \code{ExpLR}, $\gamma = \num{.3}$ & $5$ \\ 
        \midrule
        % ECG Classification  & \code{RNN}    & $150$     & $64$ \\ 
        ECGC  & \code{GRU}    & $150$     & $64$ & \code{Adam} & $\num{1e-3}$ & -- & $-1$ \\
        ECGC  & \code{LSTM}   & $150$     & $64$ & \code{Adam} & $\num{1e-3}$ & -- & $-1$ \\  
        \midrule
        CT           & \code{dtriv}  & $500$     & $48$      & \code{RMSprop} & $\num{1e-3}$ & -- & $-1$ \\
        CT           & \code{RNN}    & $10k$     & $64$      & \code{RMSprop} & $\num{5e-3}$ & \code{ExpLR}, $\gamma = \num{.85}$ & $5$ \\
        CT           & \code{GRU}    & $285$     & $48$      & \code{RMSprop} & $\num{1e-2}$ & -- & $-1$ \\
        CT           & \code{LSTM}   & $6.5k$     & $48$      & \code{RMSprop} & $\num{5e-3}$ & -- & $10$ \\
        \bottomrule 
    \end{tabular} \label{tab:params}
\end{table*}

\vspace{-3mm}
\section{\code{KANN} reproduces the latent dynamics}
\label{app:kann_quantitative}

We performed a quantitative study of the ability of $C$ to truly capture the latent dynamics. We show that indeed, \code{KANN} is able to reproduce the nonlinear dynamics of the network in Eq.(1) from the main text, to a high degree of precision, and thus we achieve the empirical justification to replace $F$ with $C$. To this end, we consider the following two metrics: \\ 1. \textbf{Relative error} of hidden states: let $\{ h_{s,t} \}$ be a collection of states over samples $s = 1,...,S$ and across time $t = 1,...,T$. We generate the predicted collection $\{ h_{s,t}^{\code{KANN}}\}$ using Eq.(6), and we compute
\begin{equation*} \label{eq:rel_err}
        e_\mathrm{rel}(\{ h_{s,t}^{\code{KANN}}\}, \{ h_{s,t} \}) = \frac{1}{T \cdot S} \sum_{s, t} |h_{s, t}^\code{KANN} - h_{s,t} |_2^2 \; / \; |h_{s,t}|_2^2 \ .
\end{equation*}
2. \textbf{Accuracy error}: let $G$ be the neural network component that takes a state and produces the output of the model, i.e., $G(h_t) = \tilde{y}_t$. We denote by $\tilde{c}_t$ the category predicted by $\tilde{y}_t$, for instance $\tilde{c}_t = \argmax(\tilde{y}_t)$. We compare the difference between $\tilde{c}_t$ and $\tilde{c}_t^\code{KANN}$, obtained from $G(h_t^\code{KANN})=\tilde{y}_t^\code{KANN}$.

We show in Fig.4 of the main text, the results of our quantitative study. For the sentiment analysis problem (Fig.4, left), we obtain $>99\%$ correspondence with the classification of the network over \emph{all} the test set, as is shown for the True Positive (TP) and True Negative (TN) columns vs. the False Positive (FP) and False Negative (FN) columns. In the ECG classification task (Fig.4, right), we reconstruct $145$ signals of the normal test set and compute their loss. There is a noticeable yet small shift in the loss histogram between the network reconstruction (blue) in comparison to our reconstruction (orange). However, the threshold for this problem set at $26$ during training (black dashed line) yields $>97\%$ agreement in classification. In particular, the false classification of normal signals (around loss $90$) appear both in the network output and in ours. Finally, we also computed the relative error of the hidden states for each of the tasks, and we show the results in Tab.~\ref{tab:rel_err}. Overall, the results demonstrate that \code{KANN} faithfully represents the latent dynamics.

\vspace{-4mm}
\begin{table} [!ht]
	\centering 
	\caption{Relative error of hidden states}
	\begin{tabular}{ccccccc} \toprule
			Task                &   Copy task   &   Sentiment analysis  &   ECG classification      \\
			\midrule
			 \#batch           &    $32$        &   $64$                &   $145$      \\
			 $e_\mathrm{rel}$  &    $0.021$     & $0.095$               & $0.0056$ \\
			\bottomrule 
	\end{tabular} 
	\label{tab:rel_err}
\end{table}

\end{document}